\documentclass{article}
\usepackage{atlas_preprint,times}

\usepackage{amsmath,amsfonts,bm}

\def\eqref#1{equation~\ref{#1}}

\def\1{\bm{1}}

\DeclareMathAlphabet{\mathsfit}{\encodingdefault}{\sfdefault}{m}{sl}
\SetMathAlphabet{\mathsfit}{bold}{\encodingdefault}{\sfdefault}{bx}{n}

\usepackage{amsmath,amssymb,mathtools,booktabs,graphicx,array,multirow,colortbl}
\usepackage{float}
\usepackage{xcolor,xspace,microtype,hyperref,url}
\definecolor{atlasblue}{HTML}{23576C}
\definecolor{atlasrow}{HTML}{EDF4F8}
\newcommand{\method}{\textsc{Atlas}\xspace}
\newcommand{\Rdom}{\mathcal R}
\newcommand{\Tdom}{\mathcal T}
\hypersetup{colorlinks=true,linkcolor=black,citecolor=black,urlcolor=black,
  pdftitle={Learn Here, Move Less Elsewhere: Input-Conditioned Plasticity from Retained-Domain Activation Atlases},
  pdfauthor={Jiangtao Lin, Bangyang Wei, Yihang Ding, Siyi Liu, Yuhan Dong}}
\graphicspath{{figures/}}
\title{Learn Here, Move Less Elsewhere:\\
Input-Conditioned Plasticity from Retained-Domain Activation Atlases}
\author{Jiangtao Lin$^{1}$ \quad Bangyang Wei$^{2}$ \quad Yihang Ding$^{1,3}$\\[0.3em]
\textbf{Siyi Liu$^{1,4}$ \quad Yuhan Dong$^{1,*}$}\\[0.8em]
\small $^{1}$Tsinghua Shenzhen International Graduate School, Tsinghua University\\
\small $^{2}$School of Vehicle and Mobility, Tsinghua University\\
\small $^{3}$SZ DJI Technology Co., Ltd.\\
\small $^{4}$Tencent Holdings Limited\\[0.3em]
\small $^{*}$Corresponding author: \texttt{dongyuhan@sz.tsinghua.edu.cn}}
\date{}

\begin{document}
\maketitle

\begin{abstract}
Task-specific fine-tuning can rewrite a language model's answers beyond the training task, complicating updates that must preserve existing behavior. We introduce \method, which turns retained-domain representations into an input-dependent rule for task adaptation. An activation atlas supplies local reference centers and directional filters to a shared low-rank residual. Target supervision learns the residual, while retained geometry shapes its action throughout training and inference. On Qwen3-8B, \method achieves lower mean retained-output Kullback--Leibler (KL) divergence than all seven published baselines at shared coding-performance requirements, with consistent advantages across multiple training seeds. Structural comparisons identify the contributions of retained reference states and directional conditioning, and answer-level analyses show fewer rewritten mathematical answers and more stable commonsense choices. Experiments spanning five backbones and two retained domains further demonstrate coding gains with reduced retained-output movement. With compact storage and modest decoding overhead, \method provides a practical mechanism for acquiring specialized skills while maintaining continuity in existing responses.
\end{abstract}

\section{Introduction}
\label{sec:intro}

A language model is trained to write code, but the resulting update also changes its answers to mathematics and commonsense questions. Shared parameters carry specialized learning into inputs that supplied no target supervision. Some original answers become incorrect, some errors are corrected, and other answers change without changing correctness. These transitions matter when applications rely on continuity across model versions. Successful adaptation therefore requires both acquiring the requested skill and controlling the reach of the update into existing behavior.

CorDA and LoRA-Null show how representative activations guide adaptation subspaces \citep{yang2024corda,tang2026loranull}. TopLoRA and SLIM make adaptation input-dependent \citep{li2025toplora,han2025slim}, while STM and TALR adjust target-token contributions \citep{wu2025stm,lin2026talr}. These methods select update subspaces or modulate adaptation without explicitly using the current token's local retained neighborhood as a geometric reference.

Retained activations occupy different neighborhoods, each offering a reference state and a pattern of local variation. To address this gap, we route each token to nearby neighborhoods and use their geometry to condition one shared task residual.

\begin{figure}[t]
  \centering
\includegraphics[width=\linewidth]{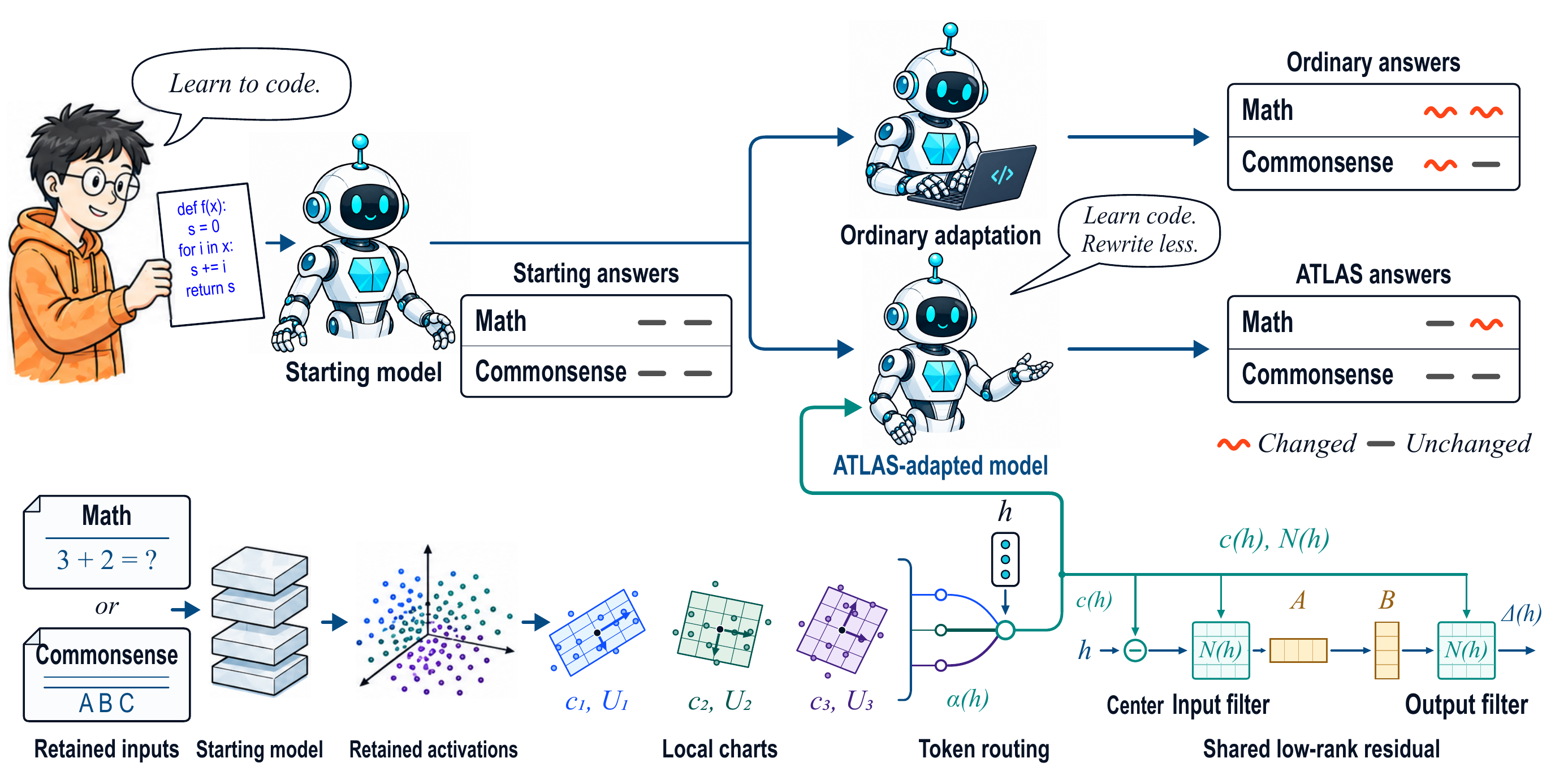}
  \caption{\textbf{Learning a task with less answer rewriting.} Both adaptation paths start from the same model. Gray marks and orange waves schematically denote unchanged and changed answers. Retained inputs illustrate domains evaluated separately. Atlas charts summarize activation neighborhoods with reference centers and local directions. The same token state $h$ drives routing and enters a shared residual conditioned on the selected geometry.}
  \label{fig:story}
\end{figure}

We realize this idea with \method, an activation atlas whose charts store centers and leading local variation (Figure~\ref{fig:story}). Target supervision trains a shared low-rank branch; the atlas shapes what it reads and emits. Because all charts share the trainable factors, richer geometric conditioning requires no additional task experts. The same rule operates during inference: \emph{supervision sets the learning destination; retained geometry gives updates a state-dependent address}.

We first evaluate \method on Qwen3-8B against seven published adaptation methods, using complete learning trajectories to compare output preservation at shared coding requirements. Experiments spanning five backbones and two retained domains extend this assessment across model scales and families. Structural comparisons then examine how centers, distances, and directions shape the residual, and answer-level analyses follow the resulting changes in mathematical answers and commonsense choices. We summarize our main contributions as follows:
\begin{itemize}
\item We propose \method, an activation-atlas method that turns local retained reference states and directions into token-dependent filters for one shared task residual. This makes retained geometry a reusable part of adaptation throughout training and inference.
\item Our method achieves lower mean retained-output KL than all seven published baselines at shared coding requirements on Qwen3-8B, with consistent advantages across three training seeds. Cross-family results extend the learning--retention benefit to GLM.
\item We connect distributional preservation to fewer rewritten mathematical answers and more stable commonsense choices. Decomposing harmful flips, incidental corrections, and within-error changes captures both correctness transitions and changes in answer identity.
\end{itemize}

\section{Related Work}
\label{sec:related}

Related methods control the effects of adaptation through learning signals, geometric constraints, or input-dependent computation.

\paragraph{Preservation through learning signals.}
Elastic Weight Consolidation penalizes changes to previously important parameters; Learning without Forgetting preserves old-network predictions on new-task inputs \citep{kirkpatrick2017ewc,li2016lwf}. GEM constrains gradients using task-wise memories, and A-GEM uses an average memory constraint \citep{lopezpaz2017gem,chaudhry2019agem}. STM excludes high-perplexity response tokens, while TALR dynamically reweights token losses \citep{wu2025stm,lin2026talr}. These approaches shape the optimization signal that learns an update. Our focus is how a shared task residual acts on different retained states, using their representations to condition its computation.

\paragraph{Geometry of preserved computation.}
Direction selection offers a geometric approach to controlling interference. Orthogonal Gradient Descent protects previous output-gradient directions; Gradient Projection Memory extracts principal activation directions for subsequent gradient projection \citep{farajtabar2020ogd,saha2021gpm}. Adapters and LoRA provide compact task parameterizations \citep{houlsby2019parameter,hu2022lora}, and O-LoRA separates low-rank task subspaces to reduce interference during language-model continual learning \citep{wang2023olora}. These studies motivate geometric constraints within compact task updates.

\paragraph{Low-rank subspace constraints.}
Within low-rank adaptation, CorDA orients weight decomposition using activation covariance, with its knowledge-preserving mode learning the smallest singular components; LoRA-Null initializes adapters in activation null spaces \citep{yang2024corda,tang2026loranull}. OPLoRA projects updates away from dominant frozen-weight singular directions, while CLoRA regularizes updates with predefined subspaces \citep{xiong2026oplora,lu2025clora}. These methods specify reference geometry for shared weights or factors, without selecting a local retained reference for each token.

\paragraph{Input-conditioned adaptation.}
Input-conditioned methods make adaptation depend on the current state. TopLoRA generates token-specific diagonal transformations within shared low-rank factors \citep{li2025toplora}. SLIM mixes LoRA experts with identity paths and adjusts identity routing using distances to target-task activation clusters \citep{han2025slim}. PASs-MoE uses pathway activation subspaces to coordinate expert routing and stabilize previously important directions in continual multimodal learning \citep{hou2026passmoe}. These methods establish token-dependent transformations and geometry-aware expert routing. ATLAS uses local retained centers and directions as explicit geometric references for the input and output of one shared residual.

We evaluate this conditioning rule against seven published baselines spanning subspace constraints, input conditioning, and token-level strategies. At shared coding requirements on Qwen3-8B, ATLAS achieves 10.16\% to 17.60\% lower mean retained-output KL (Table~\ref{tab:external}).

\paragraph{Behavioral continuity across updates.}
Prediction churn measures changes between model versions, and negative-flip analysis isolates newly introduced errors \citep{milanifard2016churn,yan2021positive}. Backward-compatible weight interpolation and MUSCLE extend compatibility objectives to language-model updates \citep{schumann2024bcwi,echterhoff2024muscle}. Our answer-level analysis follows this view by distinguishing harmful flips, incidental corrections, and changes between distinct incorrect answers. It shows fewer rewritten mathematical answers and more stable commonsense choices after code adaptation (Section~\ref{sec:behavior}).

\section{Retained Geometry for Task Adaptation}
\label{sec:method}

To make retained information part of each update, \method couples an activation atlas with a shared low-rank branch. Let $f_0$ denote the starting model, $\Tdom$ the supervised target distribution, and $\Rdom$ the retained distribution. At a selected transformer layer, a token state $h\in\mathbb R^d$ becomes $h'=h+\Delta(h)$. Target supervision trains the residual through the downstream model; the atlas supplies its input-dependent geometry.

\paragraph{Constructing the atlas.}
We first sample non-padding activations of $f_0$ on retained calibration inputs and group them into $K$ local neighborhoods. Chart $j$ stores a center $c_j\in\mathbb R^d$ and an orthonormal basis $U_j\in\mathbb R^{d\times p}$ from local principal component analysis (PCA). Its columns summarize leading activation variation around the center. The backbone and atlas remain fixed during target training; only the shared residual factors are updated. Appendix~\ref{app:protocol} specifies sampling, neighborhood construction, and insertion layers.

\paragraph{Routing a token to local geometry.}
To select the geometry relevant to the current activation, let $\mathcal J(h)$ contain the $q$ nearest centers. Soft routing assigns
\begin{equation}
  \alpha_j(h)=\frac{\exp(-\|h-c_j\|_2^2/\tau)}
  {\sum_{\ell\in\mathcal J(h)}\exp(-\|h-c_\ell\|_2^2/\tau)},
  \qquad j\in\mathcal J(h),
  \label{eq:routing}
\end{equation}
with zero weight outside this set and temperature $\tau>0$ controlling the concentration of the routing weights. The mixture defines a center and a local filter,
\begin{equation}
 c(h)=\sum_j\alpha_j(h)c_j,\qquad
 N(h)=I-\sum_j\alpha_j(h)U_jU_j^\top.
 \label{eq:filter}
\end{equation}
The filter attenuates directions represented by the local PCA bases and preserves their common orthogonal complement. Its action therefore changes with the activation neighborhood. Nonnegative weights summing to one make $N(h)$ positive semidefinite with eigenvalues in $[0,1]$. One chart for the entire retained distribution gives a global tangent-complement projector; mixing charts produces a soft attenuation operator.

\begin{figure}[t]
  \centering
\includegraphics[width=\linewidth]{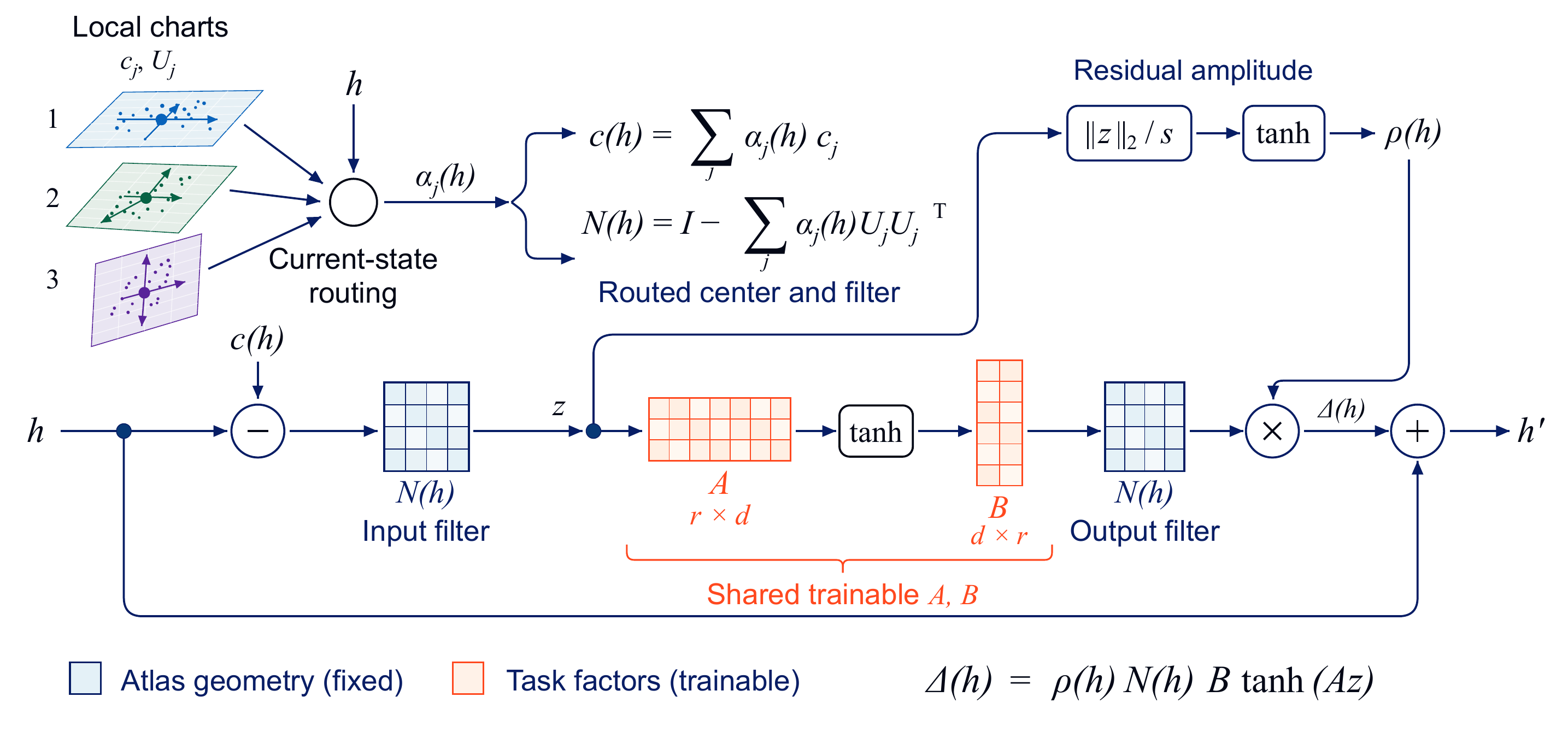}
  \caption{\textbf{Retained centers and directions condition a shared task residual.} Routing forms $c(h)$ and $N(h)$; centering and input filtering produce $z$. Shared trainable factors $A,B$ learn the task transformation. Output filtering shapes direction; $\rho(h)$ scales amplitude using $z$. The residual is added to $h$.}
  \label{fig:structure}
\end{figure}

\paragraph{Filtering the task residual.}
To limit the residual along dominant directions of retained activation variation, we apply the routed filter to both the input and output of a rank-$r$ branch:
\begin{align}
 z(h)&=N(h)\bigl(h-c(h)\bigr),
 &\rho(h)&=\tanh\!\left(\|z(h)\|_2/s\right),\label{eq:input}\\
 \Delta(h)&=\rho(h)N(h)B\tanh\!\bigl(Az(h)\bigr),
 &h'&=h+\Delta(h),\label{eq:residual}
\end{align}
where $A\in\mathbb R^{r\times d}$ and $B\in\mathbb R^{d\times r}$ are trainable. Input filtering determines which components the low-rank branch reads. Output filtering shapes the directions it emits. The positive scale $s$ controls how the residual scaling factor $\rho(h)$ grows with the filtered-state norm. Figure~\ref{fig:structure} distinguishes the atlas geometry from the learned branch.

Sharing $A$ and $B$ across charts keeps the trainable parameter count at $2dr$, independently of geometric resolution. For Qwen3-8B, $K=32$, $q=4$, $p=16$, and $r=16$ give 131,072 trainable parameters. Ordinary supervised next-token loss trains the target branch, leaving retained geometry available throughout training and inference.

We compute $N(h)v=v-\sum_{j\in\mathcal J(h)}\alpha_j(h)U_j(U_j^\top v)$ using the selected $d\times p$ bases. Input and output filtering reuse the current token's charts and routing weights, keeping the cost tied to the selected local ranks during generation.

\section{Evaluating Learning and Retained Behavior}
\label{sec:setup}

\paragraph{Tasks and models.}
We evaluate the intended use directly: learning code while maintaining existing responses on other inputs. MBPP supplies training problems \citep{austin2021mbpp}, and EvalPlus tests generated programs with the stronger MBPP+ tests \citep{liu2023evalplus}. The 224 evaluation problems are separate from the 120 training and 43 validation problems. GSM8K measures mathematics retention \citep{cobbe2021gsm8k}; CommonsenseQA (CSQA) extends evaluation to commonsense question answering \citep{talmor2019commonsenseqa}. Calibration inputs construct the atlas and supply retained examples to Replay and Output-KL. Held-out inputs measure KL and answer changes. Appendix~\ref{app:protocol} specifies data partitions, evaluation-subset overlap, and prompts.

Qwen3-8B is the main backbone for published-method and residual-structure comparisons. Qwen3.5-9B supports the detailed mathematical-answer analysis, and Qwen3-8B supplies the CSQA comparison. To examine transfer across scales and families, the full study also includes Qwen3-4B, Qwen3-14B, and GLM-4-9B \citep{yang2025qwen3,qwen2026qwen35,glm2024chatglm}. We evaluate three training seeds and complete five-epoch trajectories, using single-layer adaptation and the same evaluation protocol for starting and adapted models. Appendix~\ref{app:trajectories} presents the complete scale and retained-domain results; Appendix~\ref{app:coverage} reports target-learning assessments on other models and tasks.

\paragraph{Comparisons.}
The main comparison covers CorDA-KPM, LoRA-Null, OPLoRA ($k\in\{16,128\}$), CLoRA, TopLoRA, STM, and TALR. These methods represent directional constraints, input-conditioned transformations, and token-level training strategies. They share a single-layer adaptation budget, with native weight methods operating on their corresponding modules. To examine the role of retained information more directly, we also compare with an unconstrained residual (Plain), retained-example training (Replay), output distillation (Output-KL), and a one-chart construction (Global-PCA). Appendix~\ref{app:baselines} provides exact configurations and comparisons between residual adapters and native weight-based adapters. Validation loss selects learning rates; all three seeds and five-epoch trajectories enter the analysis.

\paragraph{Measuring learning and retained movement.}
The two objectives require complementary measurements. MBPP+ pass@1 measures code learning, while retained-output KL measures changes in next-token distributions on shared teacher-forced contexts. Selected-checkpoint tables use the lowest-validation-loss checkpoint from each curve to describe independently selected target scores.

Writing $\mathcal V$ for the evaluated context positions, the distributional measure is
\begin{equation}
 d(f,f_0)=\frac{1}{|\mathcal V|}\sum_{(x,t)\in\mathcal V}
 D_{\mathrm{KL}}\!\left(p_0(\cdot\mid x_{<t})\,\|\,p_f(\cdot\mid x_{<t})\right).
 \label{eq:kl}
\end{equation}
The reference distribution and evaluated non-padding context positions are fixed within each comparison. Squared displacement describes the emitted residual at the intervention layer; normalized answer changes measure prediction churn.

Because methods reach different coding scores during training, we compare retained movement at shared learning requirements. At requirement $k$, method $m$ contributes the smallest observed KL among checkpoints solving at least $k$ MBPP+ problems:
\begin{equation}
 D_m(k)=\min_{i:\,a_{m,i}\ge k} d_{m,i},\qquad
 \overline D_m=\frac{1}{|\mathcal K|}\sum_{k\in\mathcal K}D_m(k).
 \label{eq:comparison}
\end{equation}
Here $a_{m,i}$ is the solved-problem count, $d_{m,i}$ is retained KL, and $\mathcal K$ is the observed shared integer-score range. We geometrically average seed ratios $\overline D_{\method}/\overline D_m$. Pairwise ranges preserve learning overlap; the seven-method table uses one range common to every method and seed. This comparison makes the coding requirement explicit when assessing preservation. Appendix~\ref{app:sensitivity} reports alternative ranges, distinct scores, and single-problem sensitivity.

\section{Learning with Less Retained-Output Movement}
\label{sec:results}

\paragraph{Lower retained KL than seven published baselines.}
At the 13 shared coding requirements defined in Section~\ref{sec:setup}, \method achieves lower mean retained KL than all seven published baselines on Qwen3-8B (Table~\ref{tab:external}). The reduction ranges from 10.16\% to 17.60\% across eight configurations and holds for every training seed.

\begin{table}[!t]
  \centering
  \caption{\textbf{Lower retained KL than seven published baselines.} Qwen3-8B, GSM8K, three seeds, and 148--160 solved MBPP+ problems. Mean KL averages requirements within each seed, then summarizes seeds geometrically. All methods have 131,072 trainable parameters except TopLoRA with 135,179.}
  \label{tab:external}
  \vspace{\baselineskip}
  \setlength{\tabcolsep}{7pt}
  \setlength{\aboverulesep}{1.2pt}\setlength{\belowrulesep}{1.8pt}
  \begin{tabular}{@{}lrr@{}}
\toprule
Method & Mean KL ($\times 10^{-3}$) $\downarrow$ & ATLAS reduction (\%) \\
\midrule
\multicolumn{3}{@{}l}{\itshape Activation and directional constraints} \\
CorDA-KPM & 1.402 & 17.60 \\
LoRA-Null & 1.360 & 15.09 \\
OPLoRA ($k=16$) & 1.329 & 13.07 \\
OPLoRA ($k=128$) & 1.312 & 11.97 \\
CLoRA & 1.319 & 12.42 \\
\midrule
\multicolumn{3}{@{}l}{\itshape Input-conditioned transformation} \\
TopLoRA & 1.384 & 16.53 \\
\midrule
\multicolumn{3}{@{}l}{\itshape Token-level training signals} \\
STM & 1.285 & 10.16 \\
TALR & 1.339 & 13.75 \\
\midrule
\rowcolor{atlasrow}\multicolumn{1}{@{}>{\columncolor{atlasrow}[0pt][\tabcolsep]}l}{\textbf{ATLAS}} & \textbf{1.155} & \multicolumn{1}{>{\columncolor{atlasrow}[\tabcolsep][0pt]}r@{}}{---} \\
\bottomrule
\end{tabular}

\end{table}

The advantage spans subspace constraints, token-conditioned transformations, and token-level training strategies. Independent validation-selected comparisons with LoRA-Null and OPLoRA provide a complementary view: KL is 67.64\% lower with a mean target-score difference of $-0.10$ percentage points (Appendix~\ref{app:baselines}).

\begin{figure}[!htbp]
  \centering
  \includegraphics[width=\linewidth]{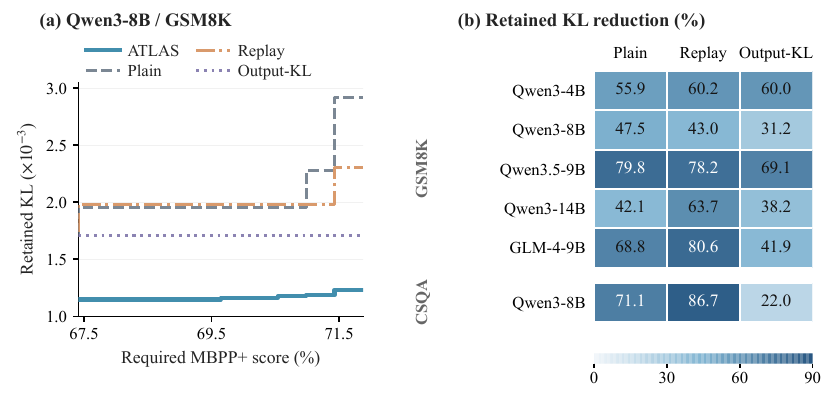}
  \caption{\textbf{Reduced retained-output movement across model settings.} (a) Qwen3-8B/GSM8K: three-seed geometric mean KL at each shared requirement of 151--161 solved MBPP+ problems. (b) Each seed uses a method pair's shared requirements; ratios of mean KL are aggregated geometrically. Appendix~\ref{app:trajectories} provides exact ranges, target scores, and trajectories.}
  \label{fig:learning}
\end{figure}

\paragraph{Learning trajectories show the preservation gain.}
To see how this benefit develops alongside code learning, Figure~\ref{fig:learning}a compares \method with Plain, Replay, and Output-KL on Qwen3-8B. The starting model solves 50.89\% of MBPP+ problems; validation-selected \method checkpoints reach 71.43\% on average. Over the shared requirements in this panel, \method consistently has lower retained KL than the three comparators. Code learning therefore accompanies a clear reduction in the update's effect on mathematics outputs.

In Figure~\ref{fig:learning}a, horizontal segments reuse checkpoints that satisfy adjacent requirements. Steps mark changes in the contributing checkpoint in at least one seed. The absolute KL axis shows how much the mathematical output distribution moves at each coding level. Complete seed-level results, equal-candidate selection, and alternative shared ranges appear in Appendices~\ref{app:baselines} and~\ref{app:sensitivity}.

\paragraph{Learning and retention across scales and families.}
To assess whether the learning--retention benefit extends beyond Qwen3-8B, we evaluate five backbones spanning 4B to 14B parameters, with GSM8K and CSQA as retained domains. Figure~\ref{fig:learning}b brings together the six evaluated settings: all 18 three-seed summaries show lower retained KL. All 54 seed-level comparisons with Plain, Replay, and Output-KL favor \method over pairwise shared coding requirements, with an aggregate KL reduction of 62.03\%. Table~\ref{tab:main} in Appendix~\ref{app:trajectories} presents every backbone's starting and validation-selected target scores alongside its retention results.

GLM-4-9B provides the cross-family test. Its starting MBPP+ accuracy is 36.61\%, and validation-selected \method checkpoints average 70.54\%. At shared coding requirements, retained KL is 67.24\% lower than Plain, Replay, and Output-KL in aggregate. This combination of substantial target learning and reduced mathematics-output movement extends the central result to another model family. Complete trajectories, geometric-resolution comparisons, and GLM answer changes appear in Appendices~\ref{app:trajectories} and~\ref{app:behavior}.

These comparisons establish improved output preservation alongside target learning. We next examine how retained reference states and local directions contribute to this benefit within the shared residual.

\section{How Retained Geometry Shapes the Residual}
\label{sec:mechanism}

\begin{table}[!htbp]
  \centering
  \caption{\textbf{Residual structures at common coding requirements.} Qwen3-8B/GSM8K, three seeds, and 151--160 solved problems. Mean KL follows Table~\ref{tab:external}'s aggregation; reductions compare \method with each row. Appendix~\ref{app:components} reports individual seeds.}
  \label{tab:mechanism}
  \vspace{\baselineskip}
  \setlength{\tabcolsep}{5pt}
  \setlength{\aboverulesep}{1.2pt}\setlength{\belowrulesep}{1.8pt}
  \begin{tabular}{@{}lrr@{}}
\toprule
Residual structure & Mean KL ($\times10^{-3}$) $\downarrow$ & ATLAS reduction (\%) \\
\midrule
\multicolumn{3}{@{}l}{\itshape Reference states and distances} \\
Residual Plain & 1.9718 & 41.30 \\
Centered residual & 1.2138 & 4.64 \\
Distance-scaled residual & 1.1849 & 2.31 \\
\midrule
\multicolumn{3}{@{}l}{\itshape Geometric conditioning} \\
PCA-derived scaling & 1.1656 & 0.70 \\
Input filter & 1.1636 & 0.52 \\
Output filter & 1.1738 & 1.39 \\
Both filters, no scaling & 1.1720 & 1.24 \\
Global-PCA & 1.1925 & 2.93 \\
\midrule
\multicolumn{3}{@{}l}{\itshape Native weight adaptation} \\
Native Weight-LoRA & 1.3302 & 12.98 \\
\midrule
\rowcolor{atlasrow}\multicolumn{1}{@{}>{\columncolor{atlasrow}[0pt][\tabcolsep]}l}{\textbf{ATLAS}} & \textbf{1.1575} & \multicolumn{1}{>{\columncolor{atlasrow}[\tabcolsep][0pt]}r@{}}{---} \\
\bottomrule
\end{tabular}

\end{table}

\paragraph{Centers establish a useful reference.}
To identify useful retained information, we compare 150 checkpoints from ten residual structures at the same coding requirements (Table~\ref{tab:mechanism}). Residual Plain and geometric variants share the insertion position, parameter budget, and paired initialization. Centered $B\tanh(Ax)$ and distance-scaled $\tanh(\|x\|_2/s_d)B\tanh(Ax)$ residuals use $x=h-c(h)$ with the same chart routing. Distance scaling matches mean amplitude on retained calibration activations. Centering reduces KL by 38.44\% relative to Plain, and distance scaling adds a 2.39\% reduction. Within this comparison, referencing retained centers supplies the largest improvement.

\paragraph{Local directions refine the conditioning.}
Full \method reduces KL by 4.64\% relative to centering and 2.31\% relative to calibrated distance scaling, with both gains present in every seed. PCA-derived amplitude improves on distance scaling by 1.62\%. The complete construction achieves the lowest aggregate KL; compact PCA-scaling and input-filtered forms retain most of its benefit. These compact forms preserve the central mechanism: retained centers, distances, and local directions condition the shared task residual. The advantage over distance scaling persists under single-problem score perturbations (Appendix~\ref{app:components}).

Table~\ref{tab:mechanism} also connects this residual comparison to native weight adaptation. Native Weight-LoRA updates the attention output projection with the same rank and parameter count. \method achieves 12.98\% lower mean KL at the common coding requirements, extending the preservation advantage to this standard low-rank parameterization.

\paragraph{Direction matters at equal effective magnitude.}
To examine residual orientation, we match each Plain residual to the effective norm of its paired \method residual. Across the four Qwen backbones, \method achieves lower KL in 11 of 12 comparisons, with an aggregate reduction of 3.89\%. The preservation benefit therefore includes a contribution from the orientation of the trained residual. Fitted local bases also outperform random bases in all 11 comparisons, supporting the value of retained geometric structure.

The local and global chart comparisons further examine geometric resolution. Qwen generally favors local charts, while GLM favors the global construction; complete per-model results appear in Appendices~\ref{app:trajectories} and~\ref{app:components}.

\paragraph{Cost of keeping geometry active.}
\method keeps retained geometry active during generation with modest decoding overhead. Appendix~\ref{app:resources} reports the computational-cost measurements.

\section{From Output Movement to Answer Rewriting}
\label{sec:behavior}

To determine how distributional preservation translates to discrete answers, we evaluate answer churn across the retained benchmark suites. We compare starting and adapted answers to each of $n$ retained questions. Let $F$ count correct-to-wrong transitions, $G$ wrong-to-correct transitions, and $W$ changes between distinct incorrect answers. For normalized answer labels, $C=F+G+W$ counts changed answers, the churn rate is $C/n$, and
\begin{equation}
 \Delta\mathrm{Accuracy}=\frac{G-F}{n}.
 \label{eq:behavior}
\end{equation}
Accuracy records the balance of errors and corrections. This decomposition follows prediction-churn and negative-flip analysis \citep{milanifard2016churn,yan2021positive}, revealing which aspects of existing behavior the adaptation preserves.

Teacher-forced KL measures distributional changes on shared reference prefixes, while answer churn records changes in normalized answer labels. In generated answers, each token also shapes the context for subsequent predictions, making answer-level evaluation useful for tracing changes through to the final response.

\begin{figure}[!htbp]
  \centering
  \includegraphics[width=\linewidth]{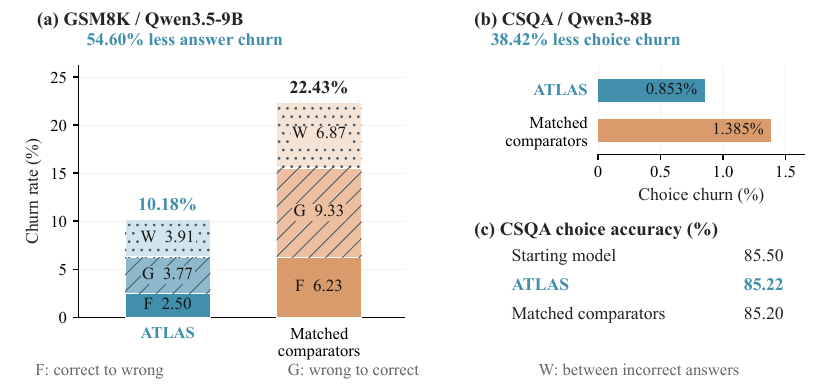}
  \caption{\textbf{Answer continuity in mathematics and commonsense.} All changes are relative to the starting model. Adapted-model values average rates equally across checkpoint pairs matched on MBPP+ counts. (a) Qwen3.5-9B: nine pairs with Plain, Replay, and Output-KL on 1,319 GSM8K questions; $F$, $G$, and $W$ share the full-question denominator. (b,c) Qwen3-8B: 12 pairs with Plain, Replay, Output-KL, and Global-PCA on 1,221 CSQA questions, showing choice churn and conditional-choice accuracy. Appendix~\ref{app:behavior} provides pairings, transition counts, and full GLM and CSQA results.}
  \label{fig:behavior}
\end{figure}

\paragraph{Fewer rewritten mathematical answers.}
The output-preservation gain carries through to mathematical answers (Figure~\ref{fig:behavior}a). On Qwen3.5-9B, the mean fraction of rewritten answers falls from 22.43\% to 10.18\%, a 54.60\% reduction across the nine matched comparisons. Correctness-transition reductions also persist under changes to batch composition and question order (Appendix~\ref{app:sensitivity}).

\paragraph{Conservation acts in both directions.}
The transition types explain what this continuity preserves. Qwen3.5 reduces harmful flips by 59.86\% and incidental corrections by 59.57\%, yielding closely balanced reductions in both transition types. The resulting continuity covers both previously correct and previously incorrect answers. Appendix~\ref{app:behavior} details the paired counts and accuracy differences, including the complete cross-family behavior analysis.

\paragraph{Stability within the incorrect-answer set.}
We also examine changes between distinct incorrect answers. In eight target-matched Qwen3-8B comparisons with LoRA-Null and OPLoRA, \method reduces these changes from 123 to 96, a 21.95\% reduction. Correctness-transition totals remain close, at 248 and 251. This comparison reveals a concrete benefit beyond correctness labels: the identity of existing answers changes less. The full transition breakdown and alternative scoring results appear in Appendix~\ref{app:behavior}.

\noindent\begin{minipage}{\linewidth}
\paragraph{More stable commonsense choices.}
Beyond mathematical answers, CSQA lets us examine which option the adapted model selects (Figure~\ref{fig:behavior}b,c). Conditional choice selects the highest-likelihood option. \method reduces choice churn by 38.42\% while maintaining nearly the same conditional-choice accuracy as the matched methods. Appendix~\ref{app:behavior} examines how these choices are expressed through short generation, reporting absolute accuracies, invalid-label rates, and format transitions together. Across mathematics and commonsense, answer-level analysis thus makes the preservation benefit concrete: fewer existing answers change after target adaptation.
\end{minipage}

\section{Conclusion}
\label{sec:conclusion}

Learning a specialized task can also change a language model's answers on other domains. We introduced \method to use retained activations when computing a shared low-rank task residual. Local centers and directional filters condition the residual at each token, while target supervision trains its shared factors. The atlas remains active throughout training and inference without introducing separate trainable branches for individual charts. On Qwen3-8B, \method achieves lower mean retained-output KL than all seven published baselines at shared coding requirements, consistently across three training seeds.

Structural comparisons identify retained centers as the largest source of improvement over the unconstrained residual, with distances and local directions further refining preservation. PCA-scaling and input-filtered variants retain most of the KL reduction observed in the structural comparison, offering simpler residual structures within the same retained-geometry formulation. Experiments across five backbones and two retained domains demonstrate coding gains with reduced output movement. Answer-level analyses connect this preservation to fewer rewritten mathematical answers and more stable commonsense choices. Separating harmful flips, incidental corrections, and changes among incorrect answers makes clear which aspects of established behavior persist through adaptation. These findings establish retained activation geometry as a compact computational resource for learning specialized skills while controlling changes to existing responses.

Future work can extend retained-geometry conditioning across layers to study where reference states most effectively guide adaptation. Coordinating layer-specific atlases would allow residual capacity to be allocated according to each layer's role in task learning and response preservation. A further direction is to select chart counts and local ranks from activation structure, using neighborhood density and directional variation to identify where finer geometry is useful. This would connect geometric resolution to the storage and computation needed to maintain it. For sequential specialization, comparing fixed and updated reference distributions could clarify how an atlas should represent both established behavior and newly acquired skills. Tracking the same questions across successive updates, with harmful flips, incidental corrections, and changes among incorrect answers recorded separately, would reveal how these reference choices shape answer continuity as the model acquires and integrates new skills.

\clearpage
\bibliography{references}
\bibliographystyle{atlas_preprint}
\clearpage
\appendix
\section{Data and evaluation protocol}
\label{app:protocol}

\subsection{Target and retained data}
The target corpus uses the sanitized MBPP training, validation, and test partitions. We train on 120 training problems and select configurations with negative log-likelihood on 43 validation problems. Code generation covers the 257 sanitized test prompts; the fixed intersection with MBPP+ contains 224 problems. All target accuracies and integer target scores in this paper use these 224 problems. MBPP training, validation, and test problems come from separate source partitions and retain their original task identifiers.

The MBPP prompt presents the task description, imports, and supplied tests, followed by a code completion. GSM8K prompts use a question followed by an answer prefix. Retained KL compares adapted and frozen distributions under the same teacher-forced sequence and averages over non-padding tokens. Output-KL regularization draws its retained training examples from the same 2,048-example pool used by Replay. GSM8K generation uses the full 1,319-problem test partition. The 512 retained test examples supply the distributional measurements, so distributional and answer-level results share these examples.

For CSQA, 2,048 training examples supply the atlas and Replay. The retained evaluation uses the 1,221-question validation partition, with a 512-question subset for KL. Conditional choice evaluates five option labels under the shared prompt. Short generation emits at most two tokens and parses an A--E label; unparsable outputs count as incorrect. Table~\ref{tab:app-data} lists the roles and overlaps.

\begin{table}[htbp]
\centering
\caption{Data roles. Atlas fitting and Replay share their training examples. Retained-KL subsets belong to the corresponding behavior evaluation set. MBPP training, validation, and test use separate source partitions.}
\label{tab:app-data}
\vspace{\baselineskip}
\begin{tabular}{llrl}
\toprule
Dataset & Use & Examples & Relation to other uses\\
\midrule
MBPP & Target training & 120 & Training partition\\
MBPP & Validation NLL & 43 & Validation partition\\
MBPP & Code generation & 257 & Test partition\\
MBPP+ & Pass@1 & 224 & Fixed intersection of test problems\\
GSM8K & Atlas fitting / Replay & 2,048 & Same training examples\\
GSM8K & Retained KL & 512 & Subset of behavior test\\
GSM8K & Answer evaluation & 1,319 & Disjoint from atlas / Replay\\
CSQA & Atlas fitting / Replay & 2,048 & Same training examples\\
CSQA & Retained KL & 512 & Subset of behavior validation\\
CSQA & Choice and generation & 1,221 & Disjoint from atlas / Replay\\
\bottomrule
\end{tabular}
\end{table}

\subsection{Training and selection}
Each main trajectory contains five epochs and uses seeds 7302026, 7302027, and 7302028. Backbone weights remain fixed. The standard ATLAS residual has rank 16 and scalar scale 64. The four Qwen studies use a learning rate of $2\times10^{-4}$. For GLM, learning-rate selection evaluates three rates with Plain and selects $4\times10^{-4}$. Replay and Output-KL use fixed coefficients within each training trajectory; Appendix~\ref{app:sensitivity} reports their coefficient sensitivity. Training uses BF16 computation, token chunking, and gradient accumulation on one RTX 5090.

The insertion module is a decoder-layer output near the middle of the backbone. Qwen3-8B and GLM-4-9B use zero-indexed layers 17 and 19. Their hidden dimension is 4,096, giving $2dr=131{,}072$ trainable residual parameters. Qwen3-4B has 81,920 residual parameters, and Qwen3-14B has 163,840. Centers and PCA bases are fixed geometric parameters and are excluded from these trainable counts.

Table~\ref{tab:app-training-config} lists the backbone identifiers and ATLAS training settings. AdamW uses $(\beta_1,\beta_2)=(0.9,0.95)$, $\epsilon=10^{-8}$, and zero weight decay. Each optimizer step accumulates 16 target examples. All five backbones use BF16 computation, gradient checkpointing, a maximum sequence length of 1,024 tokens, and five training epochs. The down-matrix $A$ is initialized with standard deviation 0.02 and the up-matrix $B$ is zero. Corresponding residual constructions share the same initial factors for each seed and begin with zero residual. The local atlas contains 32 charts of tangent rank 16, selects four charts per token, and uses routing temperature 64. The residual scaling parameter is also 64.

\begin{table}[htbp]\centering
\caption{Backbone identifiers and ATLAS training settings. Layer indices are zero-based; batch denotes target microbatch size multiplied by gradient-accumulation steps. The listed learning rates apply to all three training seeds.}
\label{tab:app-training-config}
\vspace{\baselineskip}
\begin{tabular}{lrrrrr}\toprule
Backbone identifier & Layer & Rank & Batch & Learning rate & Max. tokens\\\midrule
\texttt{Qwen/Qwen3-4B-Base} & 17 & 16 & $2\times8$ & $2\times10^{-4}$ & 1,024\\
\texttt{Qwen/Qwen3-8B-Base} & 17 & 16 & $1\times16$ & $2\times10^{-4}$ & 1,024\\
\texttt{Qwen/Qwen3.5-9B-Base} & 15 & 16 & $1\times16$ & $2\times10^{-4}$ & 1,024\\
\texttt{Qwen/Qwen3-14B-Base} & 19 & 16 & $1\times16$ & $2\times10^{-4}$ & 1,024\\
\texttt{ZhipuAI/glm-4-9b-hf} & 19 & 16 & $1\times16$ & $4\times10^{-4}$ & 1,024\\
\bottomrule\end{tabular}\end{table}

Validation-selected results choose the checkpoint with the lowest target validation NLL in each trajectory. Target-score comparisons use all five measured checkpoints. Behavioral analyses compare checkpoints at matched target scores on the same retained questions; Appendix~\ref{app:behavior-pairings} specifies each selection rule and pair. These summaries describe configuration selection, achieved learning levels, and itemwise model changes, respectively.

\subsection{Integer target-score comparison}
We apply Eq.~\ref{eq:comparison} separately within each training seed $s$, writing $D_{ms}(k)$ for method $m$'s minimum observed retained KL at integer target count $k$. Each selected value belongs to a measured checkpoint. For ATLAS and comparator $b$, the shared set $\mathcal K_{bs}$ contains integer counts in the intersection of their observed score ranges, restricted to counts above the frozen model. We report
\begin{equation}
R_{bs}
=\frac{|\mathcal K_{bs}|^{-1}\sum_{k\in\mathcal K_{bs}}D_{\mathrm{ATLAS},s}(k)}
       {|\mathcal K_{bs}|^{-1}\sum_{k\in\mathcal K_{bs}}D_{bs}(k)}.
\label{eq:app-ratio}
\end{equation}
The three-seed summary is $(R_{b1}R_{b2}R_{b3})^{1/3}$, and the relative reduction is one minus this ratio. Counts are inclusive, so $k=148,\ldots,160$ provides 13 target levels. The calculation uses the discrete score resolution directly.

The seven-method Qwen3-8B comparison uses one common set, $k=148,\ldots,160$, for all configurations and all three seeds. Pair-specific ranges retain the achieved score overlap within the core model studies. Appendix~\ref{app:sensitivity} also reports a single range shared across methods and seeds within each setting.

\paragraph{Continuous target-score averaging.}
For the supplementary continuous-target summaries, each method assigns to each of its observed target scores the smallest measured metric among checkpoints attaining at least that score. Each value is held constant from that score to the next observed score. We integrate the resulting step curve over the pair's shared observed score range, divide by its width, and form the ratio of the ATLAS and comparator averages. This calculation weights intervals by score width; the integer-target rule evaluates and averages the attainable minimum at each integer count. The continuous calculation is used for the external-baseline single-problem sensitivity analysis in Appendix~\ref{app:sensitivity}.

\subsection{Activation sampling and local PCA support}
\label{app:activation_sampling}

The standard Qwen3-8B extraction uses 2,048 GSM8K training examples and produces 4,096 vectors at decoder-layer index 17. Each example supplies two views: its formatted question prompt and that same prompt followed by its gold answer. Atlas fitting uses these retained-domain views; diagnostic MBPP examples belong to a separate split.

\begin{samepage}
Token selection is deterministic and stratified over sequence position. For zero-based example index $i$, let $b_i=i\bmod 8$. A view with $L$ non-padding tokens contributes the hidden state at zero-based non-padding rank
\[
 t_i=\left\lfloor\frac{(2b_i+1)L}{16}\right\rfloor.
\]
\end{samepage}
The two views use the same stratum and their own sequence lengths. Extraction uses left padding, a maximum length of 1,024, and bfloat16 forward computation. Atlas fitting operates on the combined 4,096 retained-domain vectors in float32 precision.

The main experiments use 32 clusters obtained with Lloyd K-means, 10 initializations, and seed 7302026. Each chart's PCA uses its assigned cluster members, centered at their mean; the leading 16 right singular vectors form its basis. Chart support therefore varies with cluster membership: chart $k$ uses $n_k$ vectors, with $\sum_k n_k=4,096$. Routing uses the four nearest chart centers and temperature 64.

The fitting-time benchmark measures an equal-support construction around each K-means center. Its support rule, $n_{\mathrm{support}}=\max(p+1,\lceil n/K\rceil)$, selects the nearest 128 vectors for basis rank $p=16$, $n=4,096$, and $K=32$. These neighborhoods may overlap. PCA is centered at the neighborhood mean, while routing retains the K-means center. Inference-throughput measurements use the cluster-member atlas from the main experiments.

\subsection{Weight-adaptation modules and method parameters}
\label{app:weight_implementation}

All weight-adaptation comparisons in Table~\ref{tab:external} act on the attention output projection (\texttt{o\_proj}) at decoder-layer index 17 of Qwen3-8B. This is a $4,096\times4,096$ matrix. The standard factors have rank 16 and scaling $\alpha/r=1$; TopLoRA uses rank 11 and $\alpha=16$ to include its token-conditioned factors within a similar parameter budget. ATLAS and residual Plain are inserted at decoder-layer output, as specified in the method section. The native Weight-LoRA calibration compares these two insertion forms.

\begin{table}[t]
\centering
\caption{Qwen3-8B weight-adaptation settings. Every method updates the same attention-output matrix. OPLoRA's protected subspace width $k$ is separate from its trainable rank. The learning rates apply to all three training seeds.}
\label{tab:weight_implementation}
\vspace{\baselineskip}

\setlength{\tabcolsep}{4pt}
\begin{tabular}{@{}lcp{0.57\linewidth}r@{}}
\toprule
Method & Rank & Mechanism parameters & Learning rate \\
\midrule
LoRA-Null & 16 & Lowest 16 activation-covariance eigenvectors for initialization & $2\!\times\!10^{-4}$ \\
OPLoRA & 16 & Weight-SVD complement projections, $k=16$ & $2\!\times\!10^{-4}$ \\
OPLoRA & 16 & Weight-SVD complement projections, $k=128$ & $2\!\times\!10^{-4}$ \\
CorDA-KPM & 16 & Smallest 16 context-oriented singular components & $4\!\times\!10^{-4}$ \\
TopLoRA & 11 & $\alpha=16$; token scale $\exp(\mathrm{RMSNorm}(Cx))$; dropout 0 & $4\!\times\!10^{-4}$ \\
STM & 16 & Retain response tokens with starting-model perplexity $\leq2.5$ & $1\!\times\!10^{-4}$ \\
CLoRA & 16 & Subspace width 512; penalty coefficient $\lambda=1$ & $4\!\times\!10^{-4}$ \\
TALR & 16 & Weight floor 0.01; temperature from the effective update batch & $2\!\times\!10^{-4}$ \\
\bottomrule
\end{tabular}
\end{table}

LoRA-Null and CorDA-KPM collect module-input activations from 2,048 retained prompt--response examples with calibration seed 7302026. Non-padding tokens enter covariance estimation. CorDA additionally normalizes each example's activation matrix by its maximum absolute entry. Its covariance regularization begins at $0.01\,\mathrm{mean}(\mathrm{diag}(C))I$ and doubles until the inverse-residual criterion is below 0.05. Both decomposition-based initializations subtract their initial $B_0A_0$ contribution, preserving the starting function.

CLoRA constructs independent input and output orthonormal bases by QR factorization, using seeds 7302026 and 7302027. Its added loss is
\[
 \frac{\lambda}{2}\bigl(\|A Q_{\mathrm{in}}\|_F^2
 +\|B^\top Q_{\mathrm{out}}\|_F^2\bigr).
\]
STM computes its response-token mask once with the starting model. TALR uses detached weights $\max\{\exp(-\ell_i/\tau),0.01\}$, where $\ell_i$ is the response-token NLL and $\tau$ is the median of the per-sequence mean response-token NLLs across the effective update batch, including accumulated microbatches. Its weighted token loss is normalized by the number of supervised response tokens.

\section{Complete learning results}
\label{app:trajectories}

The learning comparison comprises 300 Qwen checkpoints and 75 GLM checkpoints: five backbones, five methods, three seeds, and five epochs. Evaluation with CSQA retained data covers 75 Qwen3-8B checkpoints under the same five-method design. Plain's training objective is independent of the retained domain; identical trained parameters are therefore evaluated on GSM8K and CSQA.

\begin{table}[htbp]
  \centering
  \caption{\textbf{Target learning and retained-output movement across model scales and domains.} Target scores average three validation-selected checkpoints. KL ratios compare complete trajectories over pairwise shared integer target requirements. Ratios below one favor \method. All five backbones and both retained domains are included.}
  \label{tab:main}
  \vspace{\baselineskip}
  \setlength{\tabcolsep}{4pt}
  \begin{tabular}{@{}llrrrrrr@{}}
\toprule
 & & \multicolumn{3}{c}{MBPP+ (\%)} & \multicolumn{3}{c}{ATLAS / comparator KL $\downarrow$} \\
\cmidrule(lr){3-5}\cmidrule(l){6-8}
Model & Retained & Frozen & Plain & ATLAS & Plain & Replay & Output-KL \\
\midrule
Qwen3-4B & GSM8K & 2.23 & 35.71 & 14.73 & 0.441 & 0.398 & 0.400 \\
Qwen3-8B & GSM8K & 50.89 & 72.17 & 71.43 & 0.525 & 0.570 & 0.688 \\
Qwen3.5-9B & GSM8K & 62.50 & 63.54 & 65.62 & 0.202 & 0.218 & 0.309 \\
Qwen3-14B & GSM8K & 65.18 & 71.13 & 72.62 & 0.579 & 0.363 & 0.618 \\
GLM-4-9B & GSM8K & 36.61 & 65.77 & 70.54 & 0.312 & 0.194 & 0.581 \\
Qwen3-8B & CSQA & 50.89 & 72.17 & 71.73 & 0.289 & 0.133 & 0.780 \\
\bottomrule
\end{tabular}

\end{table}

Table~\ref{tab:app-seed-kl} gives all 54 primary seed-level ratios and their integer score ranges. The comparison pools 760 target levels across model--comparator--seed combinations. ATLAS has lower KL at 743 levels and lower mean KL in all 54 comparisons. The all-seed geometric reduction is 62.03\%. Absolute target accuracy, validation NLL, retained KL, and intervention-layer displacement characterize learning and preservation throughout each trajectory.

Aggregating seed ratios within each setting gives KL reductions of 58.56\% across the four Qwen backbones on GSM8K, 68.96\% on Qwen3-8B with CSQA, and 67.24\% on GLM-4-9B with GSM8K. Requiring a range shared across methods and seeds within each setting retains 53 favorable seed-level comparisons out of 54 and all 18 favorable three-seed summaries (Appendix~\ref{app:sensitivity}). The seed-level tables and complete trajectories below show the scores and KL values underlying these summaries.

\begin{table}[htbp]
\centering
\caption{Primary target-score comparisons, by training seed. Each cell gives the ratio of mean retained KL (ATLAS/comparator), followed by the inclusive integer MBPP+ range $[k_{\min},k_{\max}]$. Every value uses observed checkpoints meeting each target threshold. GSM8K is retained unless CSQA is named.}
\label{tab:app-seed-kl}
\vspace{\baselineskip}
\setlength{\tabcolsep}{3pt}
\begin{tabular}{llrrr}
\toprule
Backbone / retained domain & Comparator & 7302026 & 7302027 & 7302028\\
\midrule
Qwen3-4B & Plain & 0.444 [12,35] & 0.465 [11,37] & 0.416 [11,27] \\
Qwen3-4B & Replay & 0.343 [12,35] & 0.429 [11,37] & 0.428 [10,27] \\
Qwen3-4B & Output-KL & 0.458 [12,35] & 0.378 [8,37] & 0.370 [13,27] \\
Qwen3-8B & Plain & 0.578 [150,161] & 0.526 [144,161] & 0.477 [151,163] \\
Qwen3-8B & Replay & 0.571 [149,161] & 0.581 [148,161] & 0.558 [151,162] \\
Qwen3-8B & Output-KL & 0.621 [147,161] & 0.686 [145,161] & 0.765 [148,163] \\
Qwen3.5-9B & Plain & 0.281 [143,148] & 0.177 [141,148] & 0.165 [143,147] \\
Qwen3.5-9B & Replay & 0.177 [142,147] & 0.156 [143,148] & 0.378 [144,147] \\
Qwen3.5-9B & Output-KL & 0.280 [141,147] & 0.257 [142,149] & 0.411 [143,147] \\
Qwen3-14B & Plain & 0.558 [150,164] & 0.572 [151,164] & 0.607 [153,165] \\
Qwen3-14B & Replay & 0.287 [151,164] & 0.329 [151,166] & 0.508 [154,165] \\
Qwen3-14B & Output-KL & 0.540 [154,164] & 0.604 [151,166] & 0.724 [153,163] \\
GLM-4-9B & Plain & 0.392 [144,156] & 0.230 [141,156] & 0.336 [143,155] \\
GLM-4-9B & Replay & 0.248 [147,155] & 0.131 [144,155] & 0.225 [145,153] \\
GLM-4-9B & Output-KL & 0.635 [145,156] & 0.360 [145,154] & 0.859 [143,156] \\
Qwen3-8B / CSQA & Plain & 0.319 [150,161] & 0.260 [144,162] & 0.291 [151,162] \\
Qwen3-8B / CSQA & Replay & 0.112 [146,161] & 0.147 [142,162] & 0.142 [145,161] \\
Qwen3-8B / CSQA & Output-KL & 0.753 [147,161] & 0.781 [149,161] & 0.807 [148,161] \\
\bottomrule\end{tabular}
\end{table}

\paragraph{Target-learning trajectories.}
Across all 15 ATLAS checkpoints per Qwen backbone, mean MBPP+ scores are 9.08\% for Qwen3-4B, 69.76\% for Qwen3-8B, 64.82\% for Qwen3.5-9B, and 71.43\% for Qwen3-14B. Their frozen scores are 2.23\%, 50.89\%, 62.50\%, and 65.18\%. Mean gains are 6.85, 18.87, 2.32, and 6.25 percentage points. GLM starts at 82/224 and reaches 157, 158, and 159 solved problems at its best ATLAS checkpoint in the three seeds.

\paragraph{One-chart comparison.}
Global-PCA uses one retained-data chart and the same residual architecture. Table~\ref{tab:app-global} reports its comparison with ATLAS. The Qwen results generally favor the local atlas, while GLM favors the one-chart construction.

\begin{table}[htbp]\centering
\caption{ATLAS/Global-PCA mean-KL ratios on each pair's shared integer target range. Values below one favor the local atlas. GSM8K is retained unless CSQA is named.}
\label{tab:app-global}
\vspace{\baselineskip}
\begin{tabular}{lrrr}\toprule
Backbone / retained domain & 7302026 & 7302027 & 7302028\\\midrule
Qwen3-4B & 0.8718 & 0.8797 & 0.9278\\
Qwen3-8B & 0.9721 & 0.9746 & 0.9814\\
Qwen3.5-9B & 0.9553 & 0.9911 & 0.9593\\
Qwen3-14B & 0.9727 & 0.9797 & 1.0002\\
GLM-4-9B & 2.9425 & 3.5886 & 4.4319\\
Qwen3-8B / CSQA & 0.9751 & 0.9595 & 0.9566\\
\bottomrule\end{tabular}\end{table}

\begin{figure}[htbp]
\centering
\includegraphics[width=\linewidth]{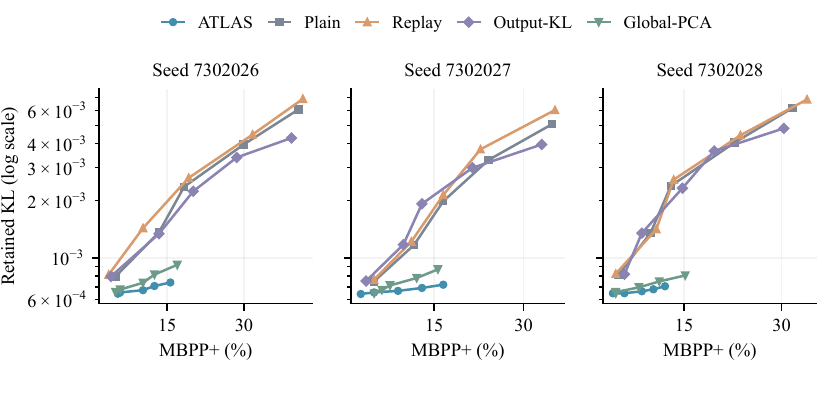}
\caption{Qwen3-4B with GSM8K retained. The three panels show individual seeds and all five measured checkpoints for each internal method. Retained KL uses a logarithmic axis.}
\label{fig:app-raw-1}
\end{figure}

\begin{figure}[htbp]
\centering
\includegraphics[width=\linewidth]{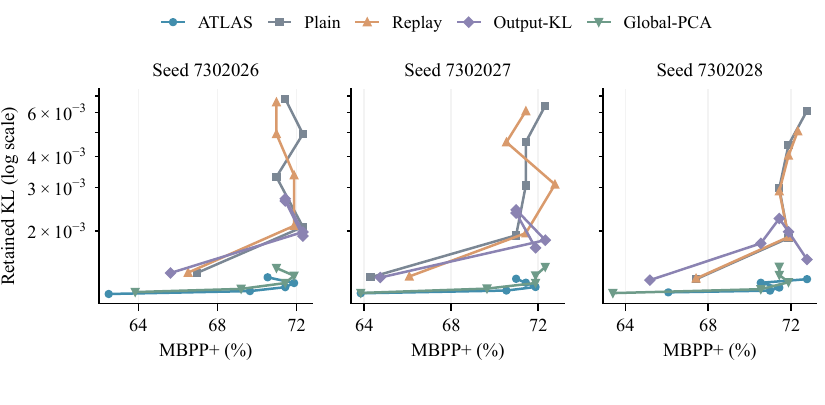}
\caption{Qwen3-8B with GSM8K retained. The three panels show individual seeds and all five measured checkpoints for each internal method. Retained KL uses a logarithmic axis.}
\label{fig:app-raw-2}
\end{figure}

\begin{figure}[htbp]
\centering
\includegraphics[width=\linewidth]{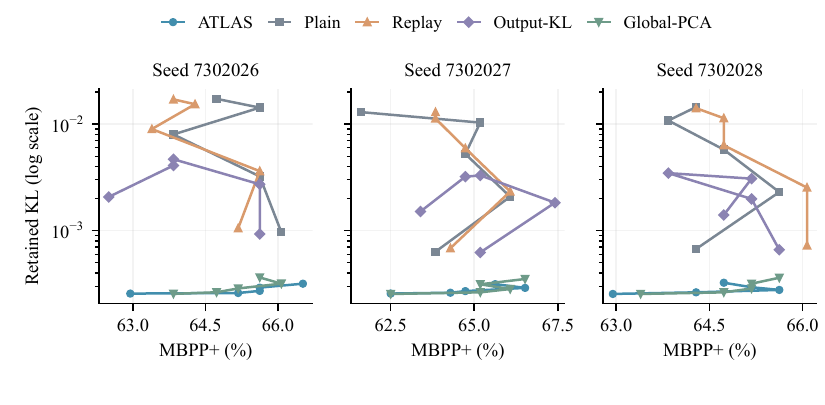}
\caption{Qwen3.5-9B with GSM8K retained. The three panels show individual seeds and all five measured checkpoints for each internal method. Retained KL uses a logarithmic axis.}
\label{fig:app-raw-3}
\end{figure}

\begin{figure}[htbp]
\centering
\includegraphics[width=\linewidth]{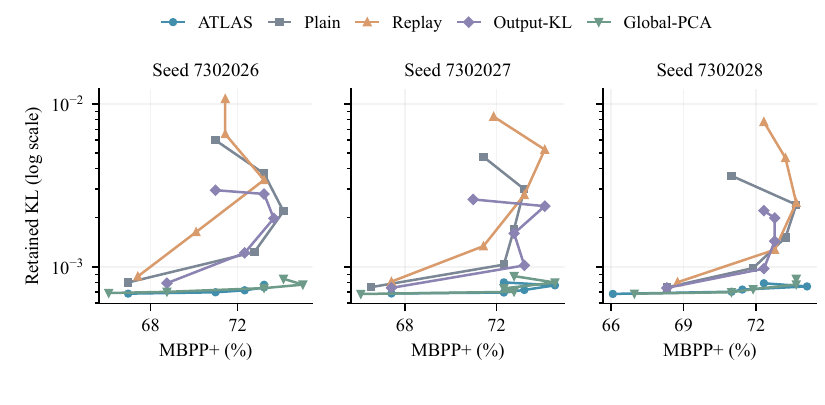}
\caption{Qwen3-14B with GSM8K retained. The three panels show individual seeds and all five measured checkpoints for each internal method. Retained KL uses a logarithmic axis.}
\label{fig:app-raw-4}
\end{figure}

\begin{figure}[htbp]
\centering
\includegraphics[width=\linewidth]{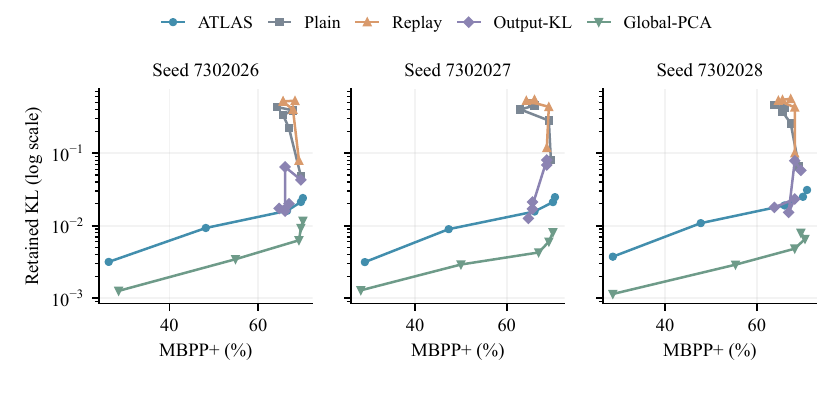}
\caption{GLM-4-9B with GSM8K retained. The three panels show individual seeds and all five measured checkpoints for each internal method. Retained KL uses a logarithmic axis.}
\label{fig:app-raw-5}
\end{figure}

\begin{figure}[htbp]
\centering
\includegraphics[width=\linewidth]{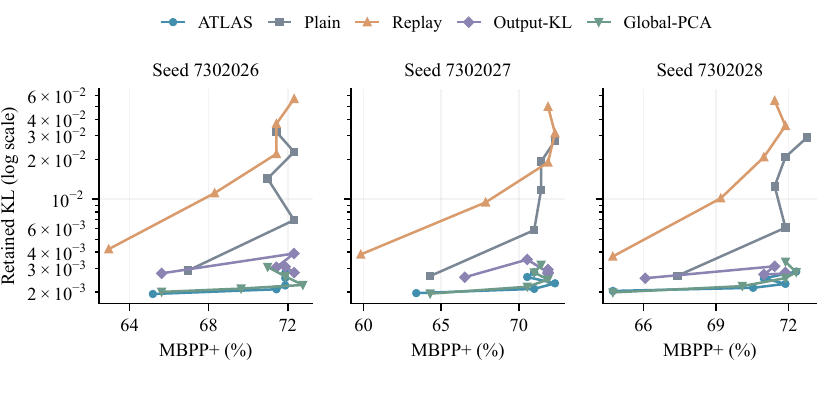}
\caption{Qwen3-8B with CSQA retained. The three panels show individual seeds and all five measured checkpoints for each internal method. Retained KL uses a logarithmic axis.}
\label{fig:app-raw-6}
\end{figure}

\paragraph{Observed checkpoints.}
Figures~\ref{fig:app-raw-1}--\ref{fig:app-raw-6} show the five-epoch measurements separately for each training seed. Each panel plots MBPP+ accuracy against retained KL at the five evaluated checkpoints per method for one training seed. The tabulated comparisons use the integer target rule in Eq.~\ref{eq:comparison}.

\clearpage
\section{Published baselines and adaptation comparisons}
\label{app:baselines}

\subsection{Implementation and tuning}
The external comparison evaluates CorDA-KPM, LoRA-Null, OPLoRA, CLoRA, TopLoRA, STM, and TALR. OPLoRA uses both $k=16$ and $k=128$. Eight configurations across three seeds give 24 training curves and 120 checkpoint evaluations on Qwen3-8B. The methods use the same MBPP training problems, validation selection, three seeds, five-epoch schedule, and approximately matched trainable parameter counts.

Weight-adapter methods retain their native linear update formulation. ATLAS and the internal Plain, Replay, and Output-KL controls use the residual-branch formulation. Standard weight and residual adapters have rank 16 and 131,072 parameters at the Qwen3-8B insertion module. TopLoRA uses rank 11 and 135,179 parameters, accommodating its token-dependent diagonal coefficients.

CorDA-KPM constructs its context-oriented decomposition from retained activations. LoRA-Null uses a retained activation subspace for initialization, and OPLoRA applies two-sided fixed weight-subspace constraints. CLoRA uses fixed random orthogonal constraints. STM masks target tokens according to frozen-model difficulty, while TALR reweights the token loss during training. TopLoRA conditions the low-rank update on each token.

\begin{table}[htbp]\centering
\caption{Learning-rate selection for the published-method comparison. Each listed method evaluates $10^{-4}$, $2\times10^{-4}$, and $4\times10^{-4}$ on seed 7302026. Minimum validation NLL over epochs 1--5 selects the rate used by all three seeds.}
\label{tab:app-learning-rates}
\vspace{\baselineskip}
\begin{tabular}{lr}\toprule
Method & Selected learning rate\\\midrule
CorDA-KPM & $4\times10^{-4}$\\
TopLoRA & $4\times10^{-4}$\\
CLoRA & $4\times10^{-4}$\\
STM & $10^{-4}$\\
TALR & $2\times10^{-4}$\\
\bottomrule\end{tabular}\end{table}

\begin{table}[htbp]\centering
\caption{External methods at the same 13 target thresholds, $k\in\{148,\ldots,160\}$ out of 224 MBPP+ problems. Entries are ATLAS/comparator ratios of mean retained KL. The last column geometrically averages the three seed ratios.}
\label{tab:app-external-seeds}
\vspace{\baselineskip}
\begin{tabular}{lrrrr}\toprule
Comparator & 7302026 & 7302027 & 7302028 & Geometric mean\\\midrule
CLoRA & 0.8893 & 0.8747 & 0.8636 & 0.8758 \\
CorDA-KPM & 0.7780 & 0.7490 & 0.9600 & 0.8240 \\
LoRA-Null & 0.8466 & 0.8566 & 0.8442 & 0.8491 \\
OPLoRA ($k=128$) & 0.8887 & 0.8768 & 0.8755 & 0.8803 \\
OPLoRA ($k=16$) & 0.8789 & 0.8623 & 0.8668 & 0.8693 \\
STM & 0.9202 & 0.8843 & 0.8912 & 0.8984 \\
TALR & 0.8803 & 0.8590 & 0.8486 & 0.8625 \\
TopLoRA & 0.7450 & 0.9230 & 0.8457 & 0.8347 \\
\bottomrule\end{tabular}\end{table}

\subsection{Independent validation selection}
To examine the learning and retention outcomes of validation-based checkpoint selection, we select each ATLAS and published-baseline trajectory independently by its lowest target-validation NLL over five epochs. Table~\ref{tab:app-validation-external} reports the resulting coding accuracy, mathematical accuracy, and retained KL for all seven published methods across their eight configurations. ATLAS achieves lower retained KL in all 24 seed-level comparisons, with geometric mean reductions of 15.47\% to 88.00\% across configurations. These results extend the retention advantage to checkpoints selected using target validation data.

TALR selects epoch three for all seeds; TopLoRA selects epochs five, five, and four for seeds 7302026, 7302027, and 7302028, respectively. ATLAS and the remaining configurations select epoch five. Within the LoRA-Null and OPLoRA subset cited in Section~\ref{sec:results}, the nine comparisons yield a geometric KL reduction of 67.64\%, a mean ATLAS-minus-comparator target difference of $-0.099$ percentage points, and a mean GSM8K difference of $-0.067$ percentage points.

\begin{table}[htbp]\centering
\caption{Independent validation selection on Qwen3-8B. Each trajectory contributes its checkpoint with the lowest target-validation NLL. MBPP+ and GSM8K are arithmetic means over three training seeds; retained KL is a geometric mean. KL reduction is one minus the geometric mean of paired ATLAS-to-comparator ratios, expressed as a percentage.}
\label{tab:app-validation-external}
\vspace{\baselineskip}
\begin{tabular}{lrrrr}\toprule
Method & MBPP+ (\%) & GSM8K (\%) & KL ($\times10^{-3}$) & Reduction (\%)\\\midrule
\rowcolor{atlasrow} ATLAS & 71.43 & 86.38 & 1.2876 & \textemdash\\
CorDA-KPM & 64.58 & 86.58 & 10.7280 & 88.00\\
LoRA-Null & 72.02 & 86.28 & 4.1632 & 69.07\\
OPLoRA ($k=16$) & 71.13 & 86.58 & 4.0335 & 68.08\\
OPLoRA ($k=128$) & 71.43 & 86.48 & 3.7502 & 65.66\\
CLoRA & 69.49 & 86.50 & 5.0130 & 74.31\\
TopLoRA & 63.69 & 86.10 & 10.6897 & 87.95\\
STM & 72.47 & 86.15 & 1.5232 & 15.47\\
TALR & 72.92 & 86.43 & 1.7447 & 26.20\\
\bottomrule\end{tabular}
\end{table}

\paragraph{Fixed target-matched checkpoints.}
The target-matched LoRA-Null/OPLoRA comparison fixes one ATLAS reference per seed and matches comparator checkpoints within one target-accuracy percentage point. Eight pairs meet this rule: three LoRA-Null, three OPLoRA-$16$, and two OPLoRA-$128$. Three pairs have equal MBPP+ counts, three differ by one problem, and two differ by two problems. All eight have lower ATLAS KL, with a geometric reduction of 50.32\%. These Qwen3-8B results summarize three training seeds.

\paragraph{Residual and weight-adapter comparison.}
ATLAS, residual Plain, and native Weight-LoRA contribute 45 Qwen3-8B checkpoints to the unified residual-structure comparison. Table~\ref{tab:mechanism} evaluates these trajectories together with the geometric variants at shared integer coding requirements. Appendix~\ref{app:residual-structure-update} reports the individual seeds over the common target range; Appendix~\ref{app:weight_implementation} specifies the insertion modules and parameter budgets.

\section{Component definitions and geometric controls}
\label{app:components}

\subsection{Residual structures at shared integer target scores}
\label{app:residual-structure-update}

The unified comparison evaluates ten Qwen3-8B residual structures over five epochs and three training seeds, giving 150 checkpoints. The structures comprise ATLAS, residual Plain, Global-PCA, four PCA component variants, centered and distance-scaled residuals, and native Weight-LoRA. Their 30 trajectories share an observed range of 151--160 solved MBPP+ problems out of 224. For each integer requirement $k$, we select the checkpoint with the lowest retained KL among those solving at least $k$ problems and average these ten KL values within each seed. Table~\ref{tab:mechanism} reports geometric means across the three seeds for both mean KL and the ratio of ATLAS KL to comparator KL, expressing the latter as a percentage reduction. Native Weight-LoRA uses the same coding and GSM8K data, rank 16, and the attention output projection at layer 17.

\begin{table}[htbp]\centering
\caption{Residual structures at the same ten integer target requirements, 151--160 solved MBPP+ problems out of 224. Each KL value averages the best achieved KL over these requirements within one training seed. Ratios divide the ATLAS value by the corresponding method value. Observed ranges describe all five checkpoints in each trajectory.}
\label{tab:app-residual-structures}
\vspace{\baselineskip}
\begin{tabular}{@{}llrrr@{}}\toprule
Structure & Seed & Observed range & Mean KL $\times10^3$ & KL ratio\\\midrule
Residual Plain & 7302026 & 150--162 & 2.0773 & 0.55920 \\
 & 7302027 & 144--162 & 2.0341 & 0.56939 \\
 & 7302028 & 151--163 & 1.8143 & 0.63536 \\
\addlinespace[2pt]
Centered residual & 7302026 & 144--163 & 1.2251 & 0.94819 \\
 & 7302027 & 147--163 & 1.2092 & 0.95781 \\
 & 7302028 & 149--162 & 1.2073 & 0.95481 \\
\addlinespace[2pt]
Distance-scaled residual & 7302026 & 136--162 & 1.1914 & 0.97499 \\
 & 7302027 & 133--163 & 1.1766 & 0.98433 \\
 & 7302028 & 141--162 & 1.1865 & 0.97150 \\
\addlinespace[2pt]
PCA-derived scaling & 7302026 & 139--160 & 1.1758 & 0.98798 \\
 & 7302027 & 143--162 & 1.1606 & 0.99796 \\
 & 7302028 & 148--163 & 1.1606 & 0.99318 \\
\addlinespace[2pt]
Input filter & 7302026 & 141--162 & 1.1615 & 1.00008 \\
 & 7302027 & 139--161 & 1.1661 & 0.99326 \\
 & 7302028 & 146--162 & 1.1632 & 0.99102 \\
\addlinespace[2pt]
Output filter & 7302026 & 141--161 & 1.1682 & 0.99437 \\
 & 7302027 & 142--163 & 1.1811 & 0.98059 \\
 & 7302028 & 144--161 & 1.1721 & 0.98343 \\
\addlinespace[2pt]
Both filters, no scaling & 7302026 & 148--162 & 1.1692 & 0.99350 \\
 & 7302027 & 148--163 & 1.1716 & 0.98855 \\
 & 7302028 & 151--162 & 1.1753 & 0.98081 \\
\addlinespace[2pt]
Global-PCA & 7302026 & 143--161 & 1.2008 & 0.96734 \\
 & 7302027 & 143--162 & 1.1961 & 0.96829 \\
 & 7302028 & 142--161 & 1.1807 & 0.97635 \\
\addlinespace[2pt]
Native Weight-LoRA & 7302026 & 141--166 & 1.3063 & 0.88923 \\
 & 7302027 & 139--163 & 1.3514 & 0.85703 \\
 & 7302028 & 144--166 & 1.3332 & 0.86464 \\
\addlinespace[2pt]
ATLAS & 7302026 & 140--161 & 1.1616 & 1.00000 \\
 & 7302027 & 143--161 & 1.1582 & 1.00000 \\
 & 7302028 & 148--163 & 1.1527 & 1.00000 \\
\bottomrule\end{tabular}\end{table}

The seed-level results support the aggregate reductions in Table~\ref{tab:mechanism}. ATLAS has lower KL in every seed for all structures except input filtering, whose 0.52\% aggregate reduction comprises two favorable seeds and one at $-0.008$\%.

The scalar used by PCA-derived scaling depends on the local atlas through $\rho(h)=\tanh(\|N(h)(h-c(h))\|_2/s)$. Its trainable branch reads the centered state, while input filtering additionally applies $N(h)$ before $A$. Centering reduces KL by 38.44\% relative to Plain; distance scaling reduces it by a further 2.39\% relative to centering; and PCA-derived scaling reduces it by 1.62\% relative to distance scaling. All three steps are favorable in every seed. Each reduction uses its own comparator denominator.

\subsection{Centering and calibrated distance scaling}
\label{app:center-distance}
To distinguish the information supplied by retained centers and distances, we compare the centered and distance-scaled residuals defined in Table~\ref{tab:app-equations}, with $x=h-c(h)$. Both use the same 32 retained centers and top-four routing with temperature 64 as ATLAS. In both constructions, $c(h)$ varies with the token and carries local retained information into the branch input. We train the layer-17 rank-16 residual for five epochs at learning rate $2\times10^{-4}$, leaving the Qwen3-8B backbone and atlas fixed. Microbatch size 1 and gradient accumulation 16 give 40 optimizer steps per curve on the 120 target-training examples. Seeds 7302026--7302028 are shared with the other structures; evaluation uses the same 224 MBPP+ problems and 512 retained examples.

We calibrate the distance scale to $s_d\approx78.6227$ by matching mean amplitude on the atlas activations. The 2,048 retained examples provide 4,096 vectors through the prompt and prompt--answer views. Calibration equates their mean $\tanh(\|x\|_2/s_d)$ to the mean PCA-derived amplitude $\tanh(\|N(h)x\|_2/64)$; both sample-average amplitudes are approximately 0.597014. PCA defines the amplitude reference during calibration; the distance-scaled residual subsequently uses centers and distances. Calibration uses retained atlas activations only, and the resulting scale is shared by all seeds. The two functions match in sample-average amplitude and retain their respective tokenwise responses.

\subsection{Sensitivity to integer target requirements and score counts}
\label{app:structure-score-sensitivity}

The comparisons with centered and distance-scaled residuals remain favorable when individual target requirements or checkpoint counts change (Table~\ref{tab:app-structure-score-sensitivity}). The analysis starts from the 151--160 requirements and keeps the measured KL values and minimum-observed-KL selection rule fixed. First, we omit each requirement in turn from all three seeds, giving ten comparisons over nine requirements. Second, we change one checkpoint's solved-problem count by $-1$ or $+1$ at a time and recompute eligibility over all ten requirements. For each control, the two methods, three seeds, five checkpoints, and two signs give 60 scenarios. In every scenario, we average selected KL within each seed and geometrically average the three ATLAS/control ratios, as in the main comparison.

\begin{table}[htbp]\centering
\caption{ATLAS retained-KL reductions (\%) under target-score sensitivity. Ranges span the stated scenarios per comparator; every scenario favors ATLAS in all three seeds.}
\label{tab:app-structure-score-sensitivity}
\vspace{\baselineskip}
\begin{tabular}{@{}lcrr@{}}\toprule
Setting & Scenarios & Centered & Distance-scaled\\\midrule
Measured counts and requirements & 1 & 4.64 & 2.31\\
Omit one target requirement & 10 & 4.44--4.79 & 2.06--2.46\\
One checkpoint count $\pm1$ & 60 & 4.33--5.08 & 2.04--2.50\\
ATLAS counts $-1$; comparator counts $+1$ & 1 & 3.62 & 1.10\\
\bottomrule\end{tabular}
\end{table}

We also simultaneously decrease all 15 ATLAS checkpoint counts by one and increase all 15 counts of the compared control by one. Under this conservative count shift, the reductions are 3.62\% and 1.10\%, with all three seeds favorable for both comparisons. Table~\ref{tab:app-structure-score-sensitivity} summarizes deterministic sensitivity ranges over these threshold omissions and hypothetical count perturbations, holding the trained checkpoints and their KL values fixed.

\subsection{Local operators}
For a token state $h$, let $x=h-c(h)$, $z=N(h)x$, and $\rho(h)=\tanh(\|z\|_2/s)$. Table~\ref{tab:app-equations} specifies the residual constructions. Centering and distance scaling use the same routed centers as ATLAS. The remaining geometric variants also use its PCA bases. PCA-derived scaling computes $z$ to obtain $\rho(h)$ while its trainable branch reads $x$; input filtering additionally inserts $N(h)$ before $A$.

\begin{table}[htbp]\centering
\caption{Residual equations. PCA-based scaled variants share $\rho(h)$; distance scaling uses $\tanh(\|x\|_2/s_d)$. The matrices $A,B$ are trainable.}
\label{tab:app-equations}
\vspace{\baselineskip}
\begin{tabular}{ll}\toprule
Variant & Emitted residual $\Delta(h)$\\\midrule
Plain & $B\tanh(Ah)$\\
Centered residual & $B\tanh(Ax)$\\
Distance-scaled residual & $\tanh(\|x\|_2/s_d)B\tanh(Ax)$\\
PCA-derived scaling & $\rho(h)B\tanh(Ax)$\\
Input filter & $\rho(h)B\tanh(AN(h)x)$\\
Output filter & $\rho(h)N(h)B\tanh(Ax)$\\
Both filters, without scaling & $N(h)B\tanh(AN(h)x)$\\
ATLAS & $\rho(h)N(h)B\tanh(AN(h)x)$\\
\bottomrule\end{tabular}\end{table}

Each $U_j$ has orthonormal columns, and the routing weights $\alpha_j(h)$ are nonnegative and sum to one. Consequently, $0\preceq\sum_j\alpha_j(h)U_jU_j^\top\preceq I$, and $N(h)$ has eigenvalues in $[0,1]$. Soft routing produces a positive-semidefinite contraction. The implementation applies $U_j(U_j^\top v)$ in token chunks, without materializing dense $d\times d$ matrices.

Local charts use K-means centers and rank-16 local PCA. The standard atlas has 32 charts and routes each token to its four nearest centers with temperature 64. Adaptation uses cluster-member PCA, as specified in Appendix~\ref{app:activation_sampling}. Global-PCA uses a single retained covariance estimate.

\paragraph{Validation-selected input filtering.}
The input-filter comparison selects ATLAS and the output-filter variant independently by minimum target-validation NLL. Displacement is the squared emitted residual normalized by hidden dimension. Adding input filtering gives 7.90\% lower KL and 40.94\% lower displacement at equal mean MBPP+ across the three seeds. Individual target-score differences are $-1.339$, $+0.446$, and $+0.893$ percentage points; all three seeds have lower retained KL with ATLAS.

\subsection{Random geometry, routing, and effective norm}
Random-local replaces fitted local directions with random directions while retaining the routing structure. Shuffled-patch disrupts the association between routed regions and their fitted bases. At matched target scores, learned charts reduce KL in 11/11 Random-local comparisons and 9/12 Shuffled-patch comparisons. Their geometric reductions are 5.53\% and 4.13\%.

On GSM8K, the effective-norm control preserves a trained Plain residual direction and rescales its emitted residual at every token to the norm produced by the paired ATLAS checkpoint. Table~\ref{tab:app-exact-norm} specifies both checkpoints in each pair across the four Qwen backbones. For Qwen3-14B seed 7302026, ATLAS is evaluated at epoch five and Plain at epoch four. Across all 12 pairs, the geometric KL reduction is 3.89\%, with 11 lower-KL comparisons. On Qwen3.5, the correctness-transition counts of ATLAS and rescaled Plain are 95/93, 79/87, and 69/71; the KL reductions are 6.15\%, 3.18\%, and 1.07\%.

\begin{table}[htbp]\centering
\caption{Twelve effective-norm comparisons on GSM8K. The Plain residual is rescaled token by token to match the effective residual norm of its paired ATLAS checkpoint. Epochs identify ATLAS/Plain. KL columns use units of $10^{-3}$; positive reduction favors ATLAS.}
\label{tab:app-exact-norm}
\vspace{\baselineskip}
\begin{tabular}{llrrrr}\toprule
Backbone & Seed & Epochs & ATLAS KL & Rescaled Plain KL & Reduction (\%)\\\midrule
Qwen3.5-9B & 7302026 & 4/2 & 0.291723 & 0.310847 & 6.152 \\
Qwen3.5-9B & 7302027 & 3/5 & 0.270298 & 0.279181 & 3.182 \\
Qwen3.5-9B & 7302028 & 2/2 & 0.263262 & 0.266117 & 1.073 \\
Qwen3-14B & 7302026 & 5/4 & 0.776070 & 0.890099 & 12.811 \\
Qwen3-14B & 7302027 & 2/1 & 0.695327 & 0.728186 & 4.512 \\
Qwen3-14B & 7302028 & 2/4 & 0.703477 & 0.738552 & 4.749 \\
Qwen3-4B & 7302026 & 1/2 & 0.645437 & 0.650675 & 0.805 \\
Qwen3-4B & 7302027 & 2/3 & 0.655505 & 0.655319 & -0.028 \\
Qwen3-4B & 7302028 & 4/2 & 0.680683 & 0.690200 & 1.379 \\
Qwen3-8B & 7302026 & 3/3 & 1.186999 & 1.228391 & 3.370 \\
Qwen3-8B & 7302027 & 4/3 & 1.237264 & 1.309790 & 5.537 \\
Qwen3-8B & 7302028 & 3/1 & 1.182595 & 1.211464 & 2.383 \\
\bottomrule\end{tabular}\end{table}

\section{Answer identity, correctness, and output format}
\label{app:behavior}

\subsection{Checkpoint pairing for the behavioral analyses}
\label{app:behavior-pairings}

Tables~\ref{tab:app-behavior-pairs-gsm} and~\ref{tab:app-behavior-pairs-additional} identify all 38 checkpoint pairs used to measure answer churn and F/G/W transitions. Subscripts $A$ and $B$ denote ATLAS and its comparator; $e$ is the training epoch and $a$ is the number of solved MBPP+ problems out of 224.

For Qwen3.5, GLM, and CSQA, selection searches the two five-epoch trajectories within each seed. It first minimizes the absolute MBPP+ score difference, then the absolute target-validation-NLL difference, then the ATLAS epoch and comparator epoch. Qwen3.5 and GLM use checkpoint identifier order as the final tie-breaker. The matching tolerance is one percentage point. All nine Qwen3.5 pairs and all twelve CSQA pairs have identical target counts. Five GLM pairs are exact ties; ATLAS solves one additional problem in the remaining four.

\begin{table}[htbp]\centering
\caption{GSM8K behavior pairings. Epochs and MBPP+ counts identify both checkpoints in each comparison. The GLM rows cover the Plain, Replay, and Output-KL comparisons summarized in Table~\ref{tab:app-glm-counts}.}
\label{tab:app-behavior-pairs-gsm}
\vspace{\baselineskip}
\begin{tabular}{@{}llrrrrr@{}}\toprule
Comparator & Seed & $e_A$ & $e_B$ & $a_A$ & $a_B$ & $a_A-a_B$\\\midrule
\multicolumn{7}{l}{\textit{Qwen3.5-9B, retained GSM8K}}\\
Plain & 7302026 & 4 & 2 & 147 & 147 & 0 \\
Replay & 7302026 & 2 & 1 & 146 & 146 & 0 \\
Output-KL & 7302026 & 3 & 1 & 147 & 147 & 0 \\
Plain & 7302027 & 3 & 3 & 145 & 145 & 0 \\
Replay & 7302027 & 2 & 1 & 144 & 144 & 0 \\
Output-KL & 7302027 & 3 & 4 & 145 & 145 & 0 \\
Plain & 7302028 & 2 & 1 & 144 & 144 & 0 \\
Replay & 7302028 & 5 & 3 & 145 & 145 & 0 \\
Output-KL & 7302028 & 3 & 1 & 147 & 147 & 0 \\
\midrule
\multicolumn{7}{l}{\textit{GLM-4-9B, retained GSM8K}}\\
Plain & 7302026 & 4 & 1 & 156 & 156 & 0 \\
Replay & 7302026 & 4 & 1 & 156 & 155 & 1 \\
Output-KL & 7302026 & 4 & 1 & 156 & 156 & 0 \\
Plain & 7302027 & 3 & 4 & 148 & 148 & 0 \\
Replay & 7302027 & 3 & 4 & 148 & 148 & 0 \\
Output-KL & 7302027 & 3 & 4 & 148 & 147 & 1 \\
Plain & 7302028 & 3 & 4 & 148 & 148 & 0 \\
Replay & 7302028 & 3 & 4 & 148 & 147 & 1 \\
Output-KL & 7302028 & 4 & 1 & 157 & 156 & 1 \\
\bottomrule\end{tabular}\end{table}

The LoRA-Null/OPLoRA comparisons hold one ATLAS reference fixed per seed, each solving 160 of 224 MBPP+ problems. The comparator epoch is selected by the smallest absolute MBPP+ gap, then the smallest validation-NLL gap, and finally the earliest epoch. All eight pairs satisfy the one-percentage-point matching window. Three have identical solved-problem counts; in the other five ATLAS solves one or two fewer problems. Both CSQA answer modalities use the same twelve checkpoint pairs.

\begin{table}[htbp]\centering
\caption{CSQA and LoRA-Null/OPLoRA behavior pairings. The CSQA choices and short generated labels come from the same checkpoints. LoRA-Null/OPLoRA comparisons use a fixed ATLAS reference per seed and a one-percentage-point matching window.}
\label{tab:app-behavior-pairs-additional}
\vspace{\baselineskip}
\begin{tabular}{@{}llrrrrr@{}}\toprule
Comparator & Seed & $e_A$ & $e_B$ & $a_A$ & $a_B$ & $a_A-a_B$\\\midrule
\multicolumn{7}{l}{\textit{Qwen3-8B, retained CSQA}}\\
Plain & 7302026 & 2 & 5 & 160 & 160 & 0 \\
Global-PCA & 7302026 & 4 & 4 & 161 & 161 & 0 \\
Replay & 7302026 & 2 & 3 & 160 & 160 & 0 \\
Output-KL & 7302026 & 5 & 4 & 161 & 161 & 0 \\
Plain & 7302027 & 2 & 2 & 159 & 159 & 0 \\
Global-PCA & 7302027 & 5 & 4 & 159 & 159 & 0 \\
Replay & 7302027 & 3 & 4 & 162 & 162 & 0 \\
Output-KL & 7302027 & 4 & 2 & 158 & 158 & 0 \\
Plain & 7302028 & 3 & 2 & 161 & 161 & 0 \\
Global-PCA & 7302028 & 3 & 3 & 161 & 161 & 0 \\
Replay & 7302028 & 1 & 1 & 145 & 145 & 0 \\
Output-KL & 7302028 & 4 & 3 & 159 & 159 & 0 \\
\midrule
\multicolumn{7}{l}{\textit{Qwen3-8B, LoRA-Null/OPLoRA comparisons}}\\
LoRA-Null & 7302026 & 3 & 5 & 160 & 160 & 0 \\
OPLoRA-16 & 7302026 & 3 & 5 & 160 & 161 & -1 \\
LoRA-Null & 7302027 & 4 & 5 & 160 & 161 & -1 \\
OPLoRA-16 & 7302027 & 4 & 2 & 160 & 162 & -2 \\
OPLoRA-128 & 7302027 & 4 & 5 & 160 & 161 & -1 \\
LoRA-Null & 7302028 & 3 & 2 & 160 & 160 & 0 \\
OPLoRA-16 & 7302028 & 3 & 3 & 160 & 160 & 0 \\
OPLoRA-128 & 7302028 & 3 & 2 & 160 & 162 & -2 \\
\bottomrule\end{tabular}\end{table}

\subsection{Definitions and aggregation}
Let $\hat y_0$ and $\hat y_m$ denote parsed frozen and adapted answers, with reference $y$. We count
\begin{align}
F &= \sum_i {\bf 1}[\hat y_{0i}=y_i,\ \hat y_{mi}\ne y_i],\\
G &= \sum_i {\bf 1}[\hat y_{0i}\ne y_i,\ \hat y_{mi}=y_i],\\
W &= \sum_i {\bf 1}[\hat y_{0i}\ne y_i,\ \hat y_{mi}\ne y_i,\ \hat y_{0i}\ne\hat y_{mi}].
\end{align}
For $n$ evaluated questions, $C=F+G+W$ counts changed normalized answers and $C/n$ is the churn rate. Correctness-transition volume is $F+G$, and accuracy changes by $(G-F)/n$. The normalization treats numerically equivalent GSM8K answer strings as the same answer.

Percentage reductions use ratios of equally weighted mean counts across matched pairs on the same 1,319 GSM8K questions. The three training seeds are the training replicates. Tables~\ref{tab:app-fg} and~\ref{tab:app-glm-counts} report the transition counts for Qwen3.5 and GLM; Appendix~\ref{app:behavior-pairings} identifies every checkpoint pair.

\subsection{Qwen3.5-9B}
The frozen model is correct on 896 questions and wrong on 423. Table~\ref{tab:app-fg} gives all nine comparisons; conditional transition rates are $F/896$ and $G/423$. Equal-weight aggregation reduces $F$ by 59.86\% and $G$ by 59.57\%. Overall churn falls from 2,663 to 1,209 events across the repeated comparison sets, a 54.60\% reduction. Appendix~\ref{app:decoding-sensitivity} reports decoding sensitivity and accuracy comparisons under both answer-scoring conventions.

\begin{table}[htbp]\centering
\caption{Qwen3.5-9B correctness transitions in the nine target-matched comparisons. Counts list ATLAS/comparator. Frozen-correct and frozen-wrong opportunities are 896 and 423 per comparison. Reductions are percentages.}
\label{tab:app-fg}
\vspace{\baselineskip}
\setlength{\tabcolsep}{3pt}
\begin{tabular}{llrrrr}\toprule
Comparator & Seed & $F$ counts & $G$ counts & $F$ reduction & $G$ reduction\\\midrule
Plain & 7302026 & 37/91 & 59/143 & 59.3 & 58.7 \\
Replay & 7302026 & 34/71 & 41/111 & 52.1 & 63.1 \\
Output-KL & 7302026 & 30/73 & 46/114 & 58.9 & 59.6 \\
Plain & 7302027 & 31/113 & 50/148 & 72.6 & 66.2 \\
Replay & 7302027 & 30/65 & 37/101 & 53.8 & 63.4 \\
Output-KL & 7302027 & 31/96 & 50/153 & 67.7 & 67.3 \\
Plain & 7302028 & 33/59 & 36/90 & 44.1 & 60.0 \\
Replay & 7302028 & 36/117 & 75/157 & 69.2 & 52.2 \\
Output-KL & 7302028 & 35/55 & 54/91 & 36.4 & 40.7 \\
\bottomrule\end{tabular}\end{table}

\subsection{GLM-4-9B}
Table~\ref{tab:app-glm-counts} pools three target-matched seed pairs per comparator, giving 3,957 question--pair observations per comparison. Accuracy differences follow from the corresponding $(G-F)/n$ values. Appendix~\ref{app:behavior-pairings} gives every paired checkpoint's coding score.

\begin{table}[htbp]\centering
\caption{GLM-4-9B answer changes relative to the frozen model. Count cells list ATLAS/comparator, pooled across three target-matched seed pairs. Accuracy differences are ATLAS minus comparator in percentage points.}
\label{tab:app-glm-counts}
\vspace{\baselineskip}
\setlength{\tabcolsep}{4pt}
\begin{tabular}{@{}lrrrrrr@{}}\toprule
Comparator & Changes & $F$ & $G$ & $W$ & \shortstack{Churn\\reduction (\%)} & \shortstack{Accuracy\\difference (pp)}\\\midrule
Plain & 515/1400 & 183/651 & 121/295 & 211/454 & 63.21 & $+7.43$\\
Replay & 515/1324 & 183/313 & 121/632 & 211/379 & 61.10 & $-9.63$\\
Output-KL & 531/759 & 193/310 & 125/190 & 213/259 & 30.04 & $+1.31$\\
\bottomrule\end{tabular}\end{table}

\subsection{External subspace methods}
Table~\ref{tab:app-external-churn} reports all transition categories for the eight Qwen3-8B LoRA-Null/OPLoRA pairs. Changes between incorrect answers decrease by 25.41\% under the answer-extraction parser and 21.95\% under numeric normalization; the table uses the latter convention throughout.

To assess variation across the evaluated questions, we draw 20,000 paired bootstrap samples of 1,319 questions. Each sample uses the same question indices across all eight checkpoint pairs and all three seeds, preserving their shared questions and ATLAS references. Percentage reductions retain equal weight for each checkpoint pair. The 2.5th and 97.5th percentiles give 95\% intervals for question-sampling uncertainty at the fixed checkpoints and training seeds.

The reduction in changes between incorrect answers has a 95\% interval of 3.03\% to 40.68\%. Its point estimates are positive in all three training seeds, at 9.09\%, 31.25\%, and 21.43\%. Table~\ref{tab:app-external-churn} presents the pooled intervals alongside the complete transition counts for each seed.

\begin{table}[htbp]\centering
\caption{Eight target-matched LoRA-Null/OPLoRA comparisons on Qwen3-8B. Panel (a) pools all pairs and reports 95\% paired-question bootstrap intervals for percentage reductions. Panel (b) shows each training seed; count cells list ATLAS/comparator. Positive reductions indicate fewer ATLAS changes.}
\label{tab:app-external-churn}
\vspace{\baselineskip}
\setlength{\tabcolsep}{4pt}
\begin{tabular}{lrrrr}\toprule
\multicolumn{5}{l}{\textit{(a) Pooled changes and question-level uncertainty}}\\
Quantity & ATLAS & Comparator & Reduction (\%) & 95\% interval (\%)\\\midrule
Total changes ($C$) & 344 & 374 & $8.02$ & $[-7.44,\,21.76]$\\
Correct-to-wrong ($F$) & 144 & 137 & $-5.11$ & $[-38.10,\,20.51]$\\
Wrong-to-correct ($G$) & 104 & 114 & $8.77$ & $[-24.27,\,37.08]$\\
Correctness transitions ($F+G$) & 248 & 251 & $1.20$ & $[-21.09,\,20.40]$\\
Changes within errors ($W$) & 96 & 123 & $21.95$ & $[3.03,\,40.68]$\\
\bottomrule\end{tabular}
\par\smallskip
\begin{tabular}{lrrrrrrr}\toprule
\multicolumn{8}{l}{\textit{(b) Changes within each training seed}}\\
Seed & Pairs & $C$ & $F$ & $G$ & $W$ & \shortstack{$C$ reduction\\(\%)} & \shortstack{$W$ reduction\\(\%)}\\\midrule
7302026 & 2 & 98/113 & 42/39 & 26/41 & 30/33 & $13.27$ & $9.09$\\
7302027 & 3 & 117/147 & 48/56 & 36/43 & 33/48 & $20.41$ & $31.25$\\
7302028 & 3 & 129/114 & 54/42 & 42/30 & 33/42 & $-13.16$ & $21.43$\\
\bottomrule\end{tabular}
\end{table}

\subsection{CSQA labels and format transitions}
The CSQA analysis uses 12 target-matched pairs, including Global-PCA, and all 1,221 questions. Conditional choice changes 125 labels for ATLAS and 203 for the comparators across pooled pairs, a 38.42\% reduction. Table~\ref{tab:csqa-full} reports absolute accuracies for conditional choice and short generation.

\begin{table}[htbp]
  \centering
  \caption{\textbf{Conditional choices and generated answer format on CSQA.} All entries are percentages. Adapted values average the 12 matched comparisons on 1,221 questions. Conditional choice selects the highest-likelihood option. Short generation requests an option letter and counts outputs outside A--E as invalid. Churn is measured relative to the starting model. The conditional-choice results are also shown in Figure~\ref{fig:behavior}b,c.}
  \label{tab:csqa-full}
  \vspace{\baselineskip}
  \setlength{\tabcolsep}{8pt}
  \begin{tabular}{@{}lrrrr@{}}
\toprule
 & \multicolumn{2}{c}{Conditional choice} & \multicolumn{2}{c}{Short generation} \\
\cmidrule(lr){2-3}\cmidrule(l){4-5}
Model state & Accuracy & Churn & Accuracy & Invalid \\
\midrule
Frozen & 85.50 & 0.00 & 65.11 & 24.49 \\
\textbf{ATLAS} & 85.22 & 0.85 & 68.16 & 21.12 \\
Matched comparators & 85.20 & 1.39 & 74.66 & 12.99 \\
\bottomrule
\end{tabular}

\end{table}

Table~\ref{tab:app-csqa-format} separates acquisition and loss of valid answer formatting. Rates average the matched comparison sets and use all evaluated questions as the denominator.

\begin{table}[htbp]\centering
\caption{CSQA short-generation format changes, as percentages of all questions. The arrows compare each adapted model with the same frozen reference.}
\label{tab:app-csqa-format}
\vspace{\baselineskip}
\setlength{\tabcolsep}{5.5pt}
\begin{tabular}{lrrrr}\toprule
Role & Invalid$\to$valid & Valid$\to$invalid & Parsed-label churn & Raw-text churn\\\midrule
ATLAS & 4.02 & 0.66 & 5.23 & 5.38\\
Matched comparators & 11.70 & 0.21 & 12.57 & 12.80\\
\bottomrule\end{tabular}\end{table}

\section{Selection and evaluation sensitivity}
\label{app:sensitivity}

\subsection{Shared target ranges}
The primary calculation uses each method pair's achieved target-score intersection. Requiring one interval shared across all primary methods and all seeds within each setting yields 53 favorable seed comparisons out of 54 and a geometric mean KL reduction of 60.92\%. All 18 three-seed model--comparator summaries remain favorable. The seven-method comparison already uses a single common interval and keeps all 24 seed comparisons favorable.

\subsection{Equal candidate counts}
The tuning sensitivity includes nine ATLAS configurations from three learning rates and three scalar scales, and three configurations each for CLoRA, STM, TALR, Output-KL, and Replay. Each configuration contributes its minimum-validation-NLL checkpoint on seed 7302026. The fixed three-candidate ATLAS subset holds scalar scale at 64 and varies learning rate. Table~\ref{tab:app-equal-candidates} applies the integer-target rule to these validation-selected configurations; the main comparison evaluates complete five-epoch trajectories at the selected configuration.

\begin{table}[htbp]\centering
\caption{Equal-candidate tuning sensitivity. Both ATLAS and each comparator contribute three validation-selected configurations. Reductions are percentages. The integer range is inclusive; TALR shares a single observed requirement with the fixed three-candidate subset.}
\label{tab:app-equal-candidates}
\vspace{\baselineskip}
\begin{tabular}{lrr}\toprule
Comparator & Integer range & KL reduction (\%)\\\midrule
CLoRA & 148--163 & 24.43\\
Output-KL & 160--163 & 38.01\\
Replay & 160--163 & 82.09\\
STM & 159--163 & 22.01\\
TALR & 163 & 20.37\\
\bottomrule\end{tabular}\end{table}

Enumerating all $\binom{9}{3}=84$ three-configuration ATLAS subsets across the five comparators gives 361 nonempty intersections. ATLAS has lower mean KL in 355, including all 314 spanning at least two integer requirements and 41 of the 47 spanning one. The three-seed fixed-configuration comparison in Table~\ref{tab:app-external-seeds} supplies the primary published-method result.

\paragraph{Retained-training strength.}
Table~\ref{tab:app-coefficient-sweep} reports the complete three-coefficient sweeps for Replay and Output-KL on Qwen3-8B with seed 7302026. Each configuration contributes its checkpoint selected by minimum target-validation NLL; all six selected checkpoints occur at epoch five. Validation NLL favors coefficient 0.3 for Output-KL and 0.1 for Replay. The Output-KL configuration at coefficient 1.0 attains a higher MBPP+ count and lower retained KL than coefficient 0.3. All three coefficients for each method enter the equal-candidate comparison in Table~\ref{tab:app-equal-candidates}.

\begin{table}[htbp]\centering
\caption{Validation-selected checkpoints from the Replay and Output-KL coefficient sweeps on Qwen3-8B, using seed 7302026. MBPP+ counts solved problems out of 224; retained KL is evaluated after checkpoint selection.}
\label{tab:app-coefficient-sweep}
\vspace{\baselineskip}
\begin{tabular}{lrrrr}\toprule
Method & Coefficient & Validation NLL & MBPP+ count & KL ($\times10^{-3}$)\\\midrule
Output-KL & 0.1 & 0.67275 & 160 & 6.1000\\
Output-KL & 0.3 & 0.67185 & 160 & 4.5893\\
Output-KL & 1.0 & 0.67487 & 163 & 1.9085\\
\midrule
Replay & 0.1 & 0.67180 & 163 & 6.6063\\
Replay & 0.3 & 0.67600 & 160 & 8.0026\\
Replay & 1.0 & 0.69211 & 163 & 23.9151\\
\bottomrule\end{tabular}
\end{table}

\paragraph{Single-problem score resolution.}
To assess score resolution, we perturb target counts by one MBPP+ problem in the continuous-target comparison. All 24 external seed directions remain favorable; the smallest retained reduction is 0.333\%. The integer analysis expresses the same test resolution directly through counts out of 224. Appendix~\ref{app:components} examines single-problem sensitivity for the residual-structure comparisons.

\subsection{Decoding batch and question order}
\label{app:decoding-sensitivity}
To assess decoding sensitivity, we evaluate the same Qwen3.5 starting model, ATLAS checkpoint, and Plain checkpoint under three conditions: the evaluation configuration, batch size eight, and a fixed shuffled question order at batch size 24. Generation remains greedy, and answer comparisons are paired by question identity.

\begin{table}[htbp]\centering
\caption{Qwen3.5 correctness transitions under decoding perturbations. Counts list ATLAS/Plain, using ATLAS epoch four and Plain epoch two on seed 7302026 in every condition.}
\label{tab:app-decoding}
\vspace{\baselineskip}
\begin{tabular}{lrrr}\toprule
Decoding condition & $F$ & $G$ & Reduction in $F+G$ (\%)\\\midrule
Evaluation batch and order & 37/91 & 59/143 & 58.97\\
Batch size eight & 34/86 & 72/153 & 55.65\\
Shuffled order, batch size 24 & 31/95 & 64/154 & 61.85\\
\bottomrule\end{tabular}\end{table}

The reduced transition volume appears under all three conditions. Across the nine primary Qwen3.5 comparisons, accuracy differences agree in direction under the GSM8K answer-extraction rule and numeric normalization: pooled ATLAS-minus-comparator differences are $-1.398$ and $-1.828$ percentage points, respectively.

\section{Target-learning engagement across models and tasks}
\label{app:coverage}

Shared target requirements depend on the learning achieved under the adaptation protocol. Additional code-learning assessments on Mistral-7B and Llama3-8B each cover five methods, three seeds, and five epochs. Across all 75 checkpoints per backbone, mean MBPP+ changes from the starting model are $-1.60$ and $-0.70$ percentage points, respectively. Gemma-2-9B has no sustained target-learning interval.

For CSQA as the target, Qwen3-8B evaluates Plain and ATLAS with one seed and five epochs on 965 questions, retaining GSM8K. From 85.803\% starting accuracy, their best gains are 0.000 and 0.104 percentage points. The shared range is 0.207 points; the protocol requires at least a one-point gain and a shared range of 0.5 points for multi-seed evaluation.

\clearpage
\section{Computational resources}
\label{app:resources}

On Qwen3-8B, ATLAS decoding takes 2.52--3.13\% longer than the frozen model; prefill takes 5.83--12.26\% longer. Table~\ref{tab:app-resources} compares the frozen model with epoch-five Plain, Global-PCA, ATLAS, and Replay checkpoints at batch sizes 1, 8, and 24. The complete local operator takes 10.37--11.14 microseconds per token in the isolated 128-token-per-example measurement, including chart selection and residual filtering.

Equal-support PCA fitting on 4,096 activation vectors from 2,048 retained examples takes 38.54 seconds, excluding activation extraction. The resulting compressed atlas file occupies 15.85 MiB. Appendix~\ref{app:activation_sampling} specifies the neighborhood construction used for fitting and inference.

\begin{table}[htbp]\centering
\caption{Qwen3-8B resource measurements on one RTX 5090. Rates count input tokens for prefill and generated tokens for decoding. Peak memory is CUDA-allocated memory in GiB. Each timed result is the median of three repetitions on the same prompt batch; decoding emits 32 tokens per example.}
\label{tab:app-resources}
\vspace{\baselineskip}
\begin{tabular}{lrrrr}\toprule
Method & Batch & Prefill (tokens/s) & Decode (tokens/s) & Peak memory (GiB)\\\midrule
Frozen & 1 & 3224.9 & 51.2 & 15.324 \\
Frozen & 8 & 5049.2 & 327.5 & 15.987 \\
Frozen & 24 & 6242.1 & 943.4 & 17.424 \\
Plain & 1 & 3215.9 & 51.5 & 15.324 \\
Plain & 8 & 5046.3 & 327.8 & 15.987 \\
Plain & 24 & 6230.6 & 942.6 & 17.426 \\
Global-PCA & 1 & 3059.3 & 49.9 & 15.340 \\
Global-PCA & 8 & 4955.5 & 320.5 & 15.987 \\
Global-PCA & 24 & 6064.1 & 922.1 & 17.426 \\
ATLAS & 1 & 3047.2 & 49.6 & 15.507 \\
ATLAS & 8 & 4602.7 & 319.5 & 15.997 \\
ATLAS & 24 & 5560.3 & 918.2 & 17.429 \\
Replay & 1 & 3221.2 & 51.4 & 15.324 \\
Replay & 8 & 5036.6 & 328.5 & 15.987 \\
Replay & 24 & 6202.4 & 950.0 & 17.426 \\
\bottomrule\end{tabular}\end{table}

\end{document}